%% file: jmlr-sample.tex
\documentclass[pmlr,twocolumn,10pt]{jmlr} 

\usepackage{booktabs}
\usepackage{siunitx}
\usepackage{enumitem}
\usepackage{multirow}
\usepackage{multicol}
\usepackage{tablefootnote}
\usepackage[normalem]{ulem}
\useunder{\uline}{\ul}{}
\usepackage[labelfont=bf]{caption}

\DeclareMathOperator*{\argmin}{argmin}

\newcommand{\equal}[1]{{\hypersetup{linkcolor=black}\thanks{#1}}}

\theorembodyfont{\upshape}
\theoremheaderfont{\scshape}
\theorempostheader{:}
\theoremsep{\newline}

\jmlrvolume{LEAVE UNSET}
\jmlryear{2022}
\jmlrsubmitted{LEAVE UNSET}
\jmlrpublished{LEAVE UNSET}
\jmlrworkshop{Conference on Health, Inference, and Learning (CHIL) 2022} 

\title[Unifying Heterogeneous EHR Systems via Text-Based Code Embedding]{Unifying Heterogeneous Electronic Health Records Systems via Text-Based Code Embedding}

 \author{%
  \Name{Kyunghoon Hur}\equal{These authors contributed equally} \Email{pacesun@kaist.ac.kr}\\
  \Name{Jiyoung Lee}\footnotemark[1] \Email{jiyounglee0523@kaist.ac.kr}\\
  \Name{Jungwoo Oh} \Email{ojw0123@kaist.ac.kr}\\
  \addr KAIST, Republic of Korea
  \AND
  \Name{Wesley Price} \Email{wjprice@mit.edu}\\
  \addr MIT, USA
  \AND
  \Name{Younghak Kim} \Email{mdyhkim@amc.seoul.kr}\\
  \addr Asan Medical Center, University of Ulsan College of Medicine, South Korea
   \AND
   \Name{Edward Choi} \Email{edwardchoi@kaist.ac.kr}\\
   \addr KAIST, Republic of Korea
 }

\begin{document}

\maketitle
\setlength{\belowcaptionskip}{-4pt}
\begin{abstract}
\input{tex/0Abstract}
\end{abstract}


\input{tex/0Datacode}

\input{tex/1Intro}
\input{tex/2RelatedWork}
\input{tex/3Methods}

\input{tex/4Results}
\input{tex/5Discussion}
\section*{Institutional Review Board (IRB)}
This research does not require IRB approval.
\acks{This work was supported by Institute of Information \& Communications Technology Planning \& Evaluation (IITP) grant (No.2019-0-00075, Artificial Intelligence Graduate School Program(KAIST)), Korea Medical Device Development Fund grant (Project Number: 1711138160, KMDF\_PR\_20200901\_0097), funded by the Korean government (the Ministry of Science and ICT, the Ministry of Trade, Industry and Energy, the Ministry of Health \& Welfare, the Ministry of Food and Drug Safety).}

\clearpage
\bibliography{jmlr-sample}
\clearpage

\appendix
\input{tex/Appendix_1}
\input{tex/Appendix_2}
\input{tex/Appendix_3}
\input{tex/Appendix_4}
\input{tex/Appendix_5}

\end{document}

%% file: tex/0Abstract.tex
Increase in the use of Electronic Health Records (EHRs) has facilitated advances in predictive healthcare. However, EHR systems lack a unified code system for representing medical concepts. Heterogeneous formats of EHR present a barrier for the training and deployment of state-of-the-art deep learning models at scale. To overcome this problem, we introduce Description-based Embedding, DescEmb, a code-agnostic description-based representation learning framework for predictive modeling on EHR. DescEmb takes advantage of the flexibility of neural language models while maintaining a neutral approach that can be combined with prior frameworks for task-specific representation learning or predictive modeling. We test our model's capacity on various experiments including prediction tasks, transfer learning and pooled learning. DescEmb shows higher performance in overall experiments compared to the code-based approach, opening the door to a text-based approach in predictive healthcare research that is not constrained by EHR structure nor special domain knowledge.

%% file: tex/0Datacode.tex
\paragraph*{Data and Code Availability}
This paper uses MIMIC-III and eICU,
which are publicly available on the PhysioNet repository \citep{mimiciii, eicu}. More
details about datasets can be found at Section 3.1.
Our code implementation is available is available on Github.\footnote{\url{https://github.com/hoon9405/DescEmb}}

%% file: tex/1Intro.tex
\section{Introduction}
\label{sec:intro}

Increased adoption of electronic health record (EHR) systems offers great potential for EHR-based predictive models to improve healthcare quality. Deep learning models have shown comparable or better performance in diagnosing or predicting various medical events. \citep{lipton2015learning, gulshan2016development, rank2020deep}. 

\begin{figure*}[ht]
  \centering
  \includegraphics[width=\linewidth]{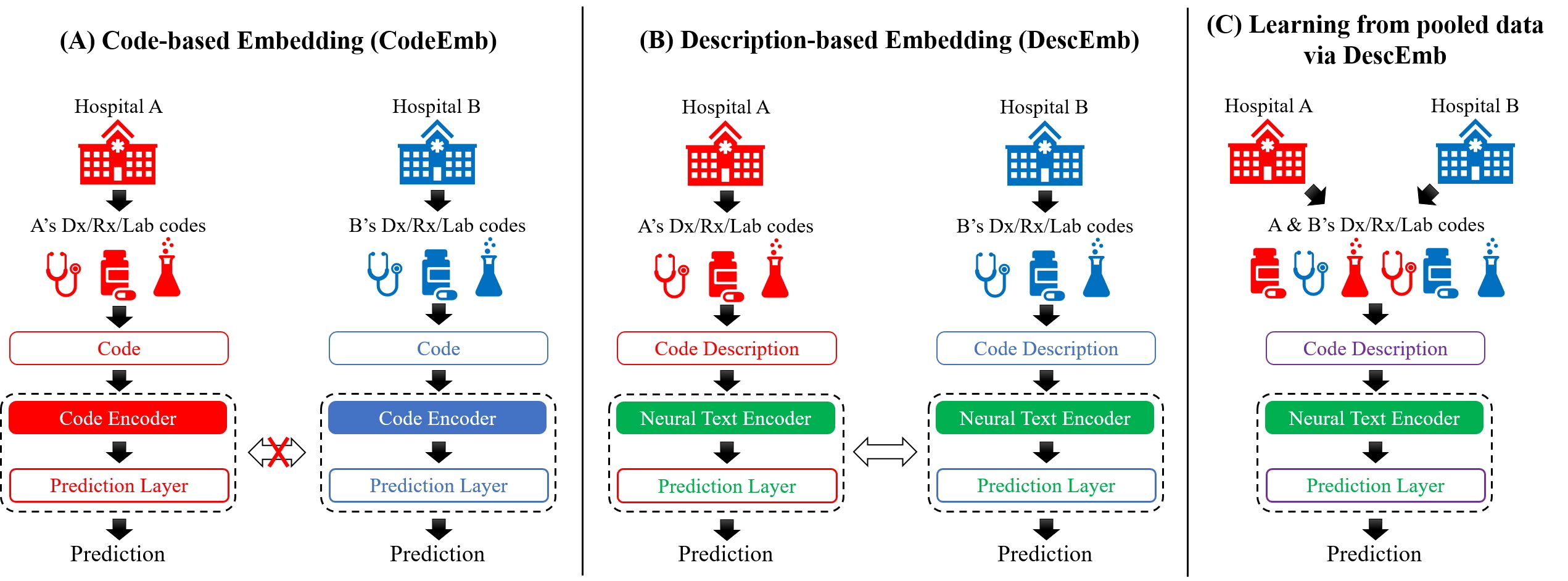}
  \caption{\label{Overview} \textbf{CodeEmb and DescEmb concept visualization.} (A) CodeEmb: predictive models are trained with code-base embedding. The code encoders and the prediction layers cannot be shared among different hospitals due to heterogeneity of the code systems. (B) DescEmb: predictive models are trained on description-based embeddings derived from the text encoder. Due to the code-agnostic nature of the text-encoder, both the text encoders and the prediction layers can be transferred between different hospitals, unlike (A). (C) Learning from pooled data via DescEmb: we can pool heterogeneous hospital data into one dataset and train jointly, thus increasing the deployment efficiency.}
\end{figure*}

However, the heterogeneity of EHR systems among hospitals presents barriers for applications of EHR-based deep learning models. Contemporary EHRs rely on data systems ranging from standardized codes (e.g. ICD9, LOINC) to free-text entry. Therefore, modern deep learning approaches are based on learning the representations of these codes, an approach we refer to as `code-based embedding'. However, this paradigm does not allow a model to be transferred from one environment to another nor to be trained on large EHR data collected from multiple hospitals that use heterogeneous EHR formats. Consequently, modern deep learning prediction models are missing out on the opportunity to be scaled up. This challenge could be alleviated by mapping codes from one system to another, or by converting all EHR data to Common Data Model (e.g. OMOP, FHIR) \citep{rajkomar2018scalable}. However, this requires significant human effort and domain knowledge and may not even be possible, depending on the code system at hand.

In this paper, we suggest code-agnostic text-based representation learning. Since each medical code has a text description that represents its semantic property, we propose Description-based Embedding, DescEmb. DescEmb adopts a neural text encoder to convert medical codes to contextualized embeddings, allowing us to map medical codes of different formats to the same text embedding space. Figure \ref{Overview} gives a visual summary of our model framework; instead of directly embedding the medical codes as in (A), the prediction layer takes a series of vectors representing code descriptions passed through a neural text encoder as in (B) and (C). Our principled approach yields improved predictive performance compared to the Code-based Embedding, CodeEmb, and makes it possible to train models on differently formatted EHR data interchangeably due to its code-agnostic nature.

We test our framework on two EHR datasets, the Medical Information Mart for Intensive Care (MIMIC-III) \citep{mimiciii} and eICU Collaborative Research Database \citep{eicu}, which use completely different medical code systems. Based on extensive experiments using five prediction tasks under diverse settings (e.g. standard single-domain learning, zero-shot and few-shot transfer learning, pooled learning), the best model of DescEmb demonstrates superior or comparable performance to the best model of CodeEmb in the vast majority of cases, outperforming by an average of 2.6\%P AUPRC. 

The main contributions of our work can be summarized as follows:

\begin{enumerate}[leftmargin=6.5mm]
\item[1.] DescEmb achieves comparable or superior performance to CodeEmb on a comprehensive set of common clinical predictive tasks. Detailed results can be found in Table \ref{eicuAUPRC}, \ref{mimicauprc}.
    \item[2.] Two differently structured EHR can be used to train and test predictive models interchangeably while rarely sacrificing model performance, often showing higher performance than when training on a single EHR. Visualized results can be found in Figure \ref{fewshot}.
    \item[3.] Two differently structured EHR can be pooled into one dataset and trained jointly with a description-based representation without the need for additional preprocessing or domain knowledge. For the result, refer to Table \ref{pooled}.
    \item[4.] DescEmb shows notable performance in overall experiments, opening the door to a text-based approach in predictive healthcare research that is not constrained by EHR structure nor special domain knowledge.

\end{enumerate}

%% file: tex/2RelatedWork.tex
\section{Related Work}
\subsection{Neural Text Encoders}
Early text embedders encode each word in a vocabulary as a vector whose semantic similarity to other words is represented by a distance measure (e.g. cosine similarity), or distribution of words \citep{mikolov2013efficient, pennington2014glove}. Recently, Bidirectional Encoder Representations from Transformers (BERT) and its variants \citep{bert, lan2019albert, liu2019roberta, yang2019xlnet} have shown improvements on various tasks in natural language processing (NLP). They employ a pre-training strategy, Masked Language Modeling (MLM) and Next Sentence Prediction (NSP), to learn contextual text representations that incorporates the complex relationships within the input text.

In the biomedical domain, several studies have developed BERT variants further trained on medical or clinical corpora. These studies have continually pre-trained their models on research articles from PubMed \citep{biobert}, MIMIC-III clinical text \citep{clinicalbert}, or a combination of the two \citep{bluebert}, and scratch trained on articles from PubMed \citep{pubmedbert}. 

\subsection{Representation Learning for Predictive Healthcare}
Predictive models with EHR data use various architectures such as autoencoders \citep{miotto2016deep, che2015deep} and recurrent neural networks (RNN) \citep{lipton2015learning, choi2016doctor, choi2016retain}. Other model architectures are also used for predictive healthcare such as gradient boosted machines \citep{chen2019deep}, convolutional nets \citep{nguyen2016deep, landi2020deep}, and Transformer-based models \citep{song2018attend, shang2019pre, choi2020learning}.

Previous research approaches are focused on code-based embedding. Our paper deals with the unification of heterogeneous code systems in EHR—and therefore sits independent to these previous works. As such, our proposed approach can be combined with prior frameworks.

%% file: tex/3Methods.tex
\section{Methods}
\subsection{Datasets}

\begin{figure*}[ht]
  \centering
  \includegraphics[width=0.7\linewidth]{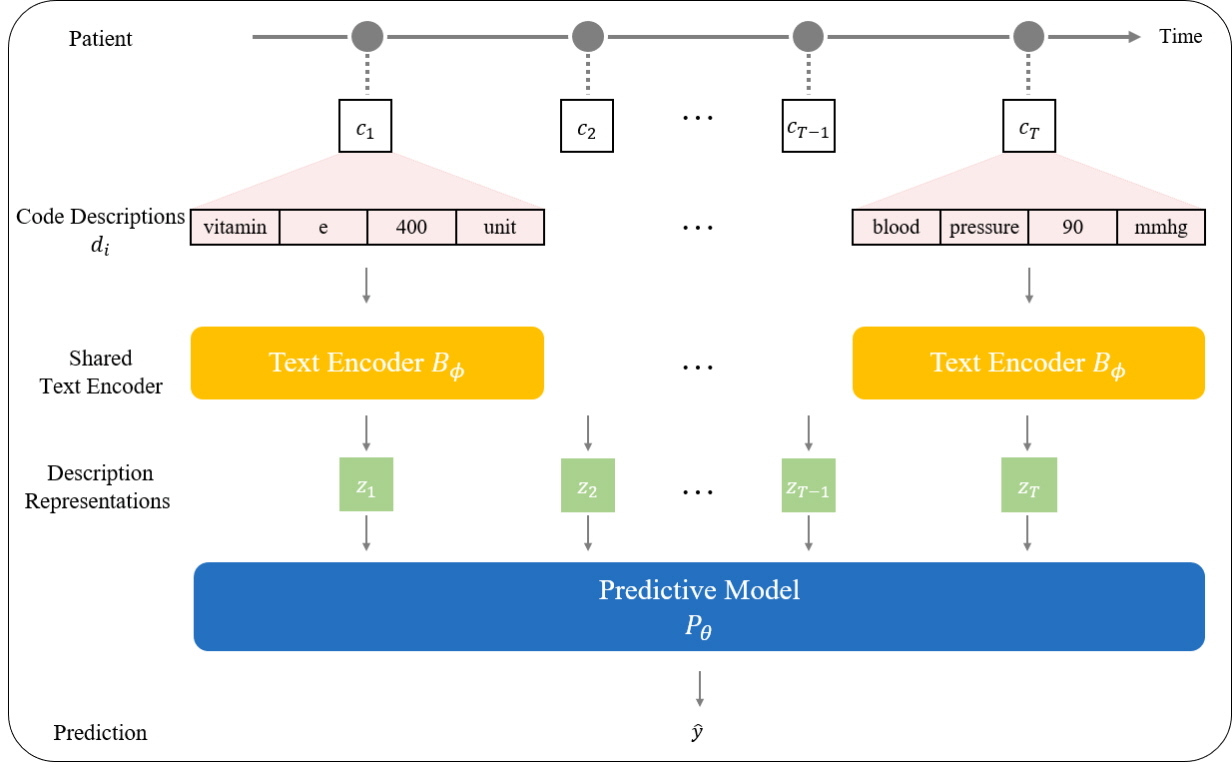}
  \caption{\label{Model} \textbf{DescEmb model framework.} On the top, the patient timeline from the ICU admission is represented as a line. Each dot on the line is a code $c_{i}$ which can be any medical event. Each code $c_{i}$ can be converted to its own description $d_{i}$. The neural text encoder $\textit{B}_{\phi}$ accepts the description $d_{i}$ and produces its latent representation $\textbf{\textit{z}}_{i}$. Given all $\textbf{\textit{z}}_{1}, \textbf{\textit{z}}_{2}, \ldots, \textbf{\textit{z}}_{T}$, the predictive model $\textit{P}_{\theta}$ predicts the outcome $\hat{y}$.}
\end{figure*}

We draw on two large, publicly available datasets: the Medical Information Mart for Intensive Care III (MIMIC-III) \citep{mimiciii}, and the eICU Collaborative Research Database (eICU) \citep{eicu}. MIMIC-III includes all patients admitted to the intensive care unit (ICU) at Beth Israel Deaconess Medical Center from 2001 to 2012, and contains over 60,000 unique ICU stays with millions of observations. The eICU Collaborative Research Database is a multi-center database comprised of de-identified health data associated with over 200,000 ICU stays across the United States between 2014-2015.

Both MIMIC-III and eICU contain time-stamped records of medical events such as labs, medications, and drug inputs for each patient stay. MIMIC-III and eICU are recorded based on completely different code structures throughout the data. For example, the clinical concept “an infusion event of nitroglycerin” is represented in eICU as the string “Nitroglycerin (mcg/min)”. However, the same semantic concept would be represented in MIMIC-III using the in-house item ID 222056 (a Metavision code, for "Nitroglycerin"); item ID 30049 (a CareVue code, for "Nitroglycerin"); or item ID 30121 (a CareVue code, for "Nitroglycerin-k"). The same goes for all medical events including diagnosis, medications, labs, etc. Consequently, we aggregate descriptions and values for comparability between formats and do not perform any within-code string manipulation. Detailed data preprocessing is provided in Appendix~\ref{supp:preprocess}.

\begin{figure*}[ht]
  \centering
  \includegraphics[width=0.8\linewidth]{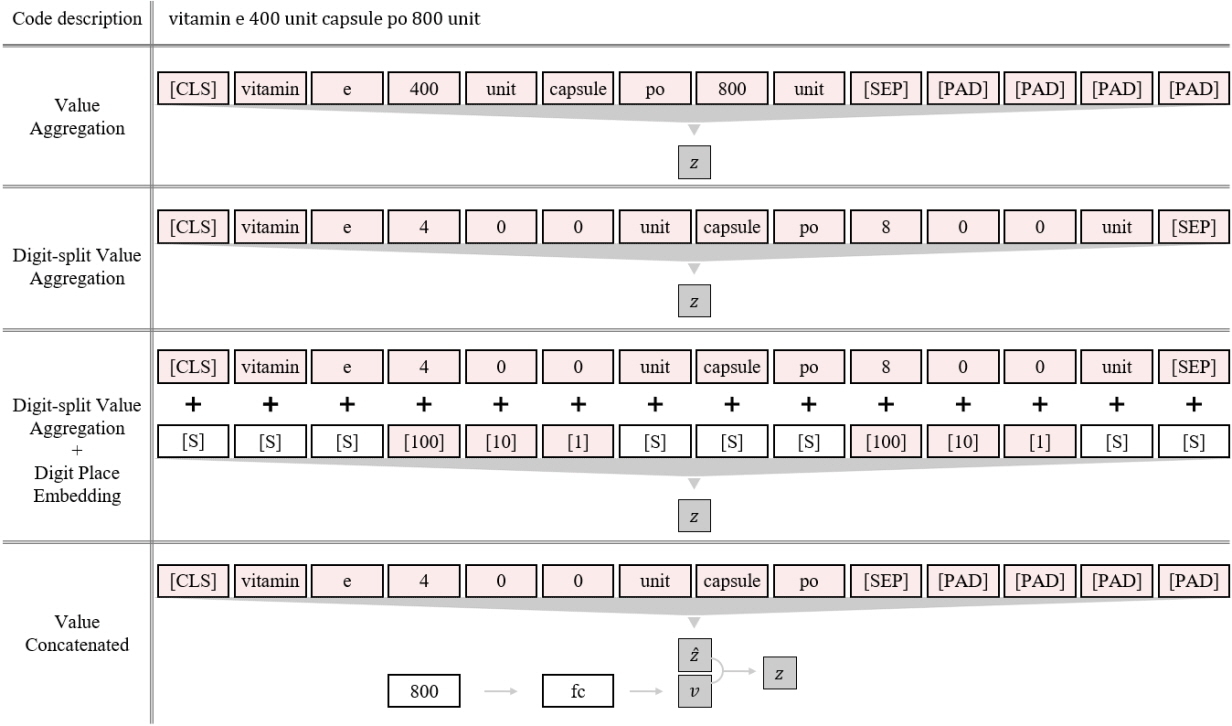}
  \caption{\label{value} \textbf{Various methods of incorporating numeric values.} We introduce four value embedding methods to represent both the code descriptions and the associated numeric values (e.g. “10” in “Tylenol 10 tabs”).}
\end{figure*}

\subsection{Structure of Electronic Health Records}
In this section, we describe the structure of EHR and introduce the notations to be used throughout the paper. Let $p^{i}$ denote the \textit{i-th} patient in the EHR data. As our problem setting is focused on individual patients, we drop the superscript when clear. A single patient $p$ can be seen as a series of medical events $(c_{1}, c_{2}, \ldots, c_{T})$ for $c_{i} \in \mathcal{C}$ where $\mathcal{C}$ denotes the set of all medical events such as diagnoses or prescriptions. Each event $c_{i}$ is typically timestamped, giving us the sequence of time information $(t_{1}, t_{2}, \ldots, t_{T})$. 

A single medical event $c_{i}$ is often associated with a text description. For example, if $c_{i}$ were a prescription event, it could be accompanied by the medication name (e.g. “Aspirin 300mg Tab.”). If it were a diagnosis event, it could come with an ICD-9 code (e.g. 401.9), which in turn has its own description (“Unspecified essential hypertension”). We use $d_{i}$ to denote this text description, which consists of a sequence of words (or sub-words) ($w_{i,1}$, $w_{i,2}$ \ldots, $w_{i,n}$) for $w_{i,j} \in \mathcal{W}$ where $\mathcal{W}$ is the entire vocabulary.

Typically, two different medical institutions employ different $\mathcal{C}$’s, such as when one hospital uses ICD-9 diagnosis codes while another uses SNOMED diagnosis codes. The vocabulary $\mathcal{W}$, however, is the same for all hospitals as long as they use the same language. We propose DescEmb, a new framework for predictive healthcare, based on this observation.

\subsection{Model Architecture}

Previous deep learning predictive models for EHR data typically have an embedding layer (or a lookup table) $\textit{E}_{\psi}$ with trainable parameters $\psi$, which converts a single medical event $c_{i}$ to its corresponding vector representation $\textbf{\textit{c}}_{i} \in \mathbb{R}^a$ where $a$ is the dimension size. 
Instead of directly converting $c_1, \ldots, c_T$ to $\textbf{\textit{c}}_1, \ldots, \textbf{\textit{c}}_T$ with a trainable lookup table, DescEmb derives the latent representation of $c_{i}$, denoted as $\textbf{\textit{z}}_{i}$, based on its text description $d_{i}$. We feed $d_{i}$ to the shared text encoder, $\textit{B}_{\phi}$, to obtain the description representations, $\textbf{\textit{z}}_{i} \in \mathbb{R}^{b}$ where $b$ is the output dimension. Repeating this for all events in the given patient $p$, we can obtain a sequence of contextualized medical event representations $(\textbf{\textit{z}}_{1}, \textbf{\textit{z}}_{2}, \ldots, \textbf{\textit{z}}_{T})$, which in turn is given to the prediction layer $\textit{P}_{\theta}$ (e.g. RNN) with trainable parameters $\theta$ to make a prediction $\hat{y}$ (Fig \ref{Model}.) The entire process of DescEmb can be summarized as below, with comparison to CodeEmb.
\begin{flalign}
& \mbox{Given a patient record $p = (c_1, c_2, \ldots, c_T$),} \nonumber && \\
& \quad \mbox{\textit{Code-based Embedding}:} \nonumber && \\
& \qquad \quad \textbf{\textit{c}}_i = \textit{E}_\psi(c_i) \nonumber && \\
& \qquad \quad \hat{y} = \textit{P}_{\theta}(\textbf{\textit{c}}_1, \textbf{\textit{c}}_2, \ldots, \textbf{\textit{c}}_T) && \\
& \quad \mbox{\textit{Description-based Embedding}:} \nonumber && \\
& \qquad \quad d_i = (w_{i,1}, w_{i,2}, \ldots, w_{i,n}) && \nonumber \\
& \qquad \quad \textbf{\textit{z}}_i = \textit{B}_{\phi}(d_i) && \\
& \qquad \quad \hat{y} = \textit{P}_{\theta}(\textbf{\textit{z}}_1, \textbf{\textit{z}}_2 \ldots, \textbf{\textit{z}}_T) &&  \nonumber
\end{flalign}

\subsection{Text Encoder}
The text encoder $\textit{B}_{\phi}$ in DescEmb can be any model that can generate a representation $\textbf{\textit{z}}_{i}$ from a given description $d_{i}$. We tested two model architectures for the text encoder: Bi-Directional Recurrent Neural Networks (Bi-RNN) and Bidirectional Encoder Representations from Transformers (BERT).  For Bi-RNN, we derived the $\textbf{\textit{z}}_{i}$ by concatenating the last hidden states from each direction. For BERT, we used the output vector from the [CLS] token as $\textbf{\textit{z}}_{i}$. We conducted experiments on different sizes of models that are pre-trained on a massive amount of general text such as Bert-tiny (2-layers), Bert-mini (4-layers), Bert-small (4-layers), Bert-base (12-layers). 

We also conducted experiments on BERTs that are pre-trained on clinical text, such as BioBERT \citep{biobert}, ClinicalBERT \citep{clinicalbert}, and BlueBERT \citep{bluebert}. We compared these models with BERTs that are pre-trained on the general domain. (results can be found in Appendix~\ref{supp:domain_res}). Moreover, we further pre-trained the text encoder on our dataset using Masked Language Modeling (MLM), following the original BERT procedure, to better fit the text encoder to our dataset. We did not include values during MLM since predicting values from descriptions is meaningless considering the various patient statuses.
\vspace{-3mm}
\subsection{Value Embedding}
\vspace{-2mm}
In the context of drug prescriptions, dosage or rate of infusion can be useful information to represent the patient's status. 
Hence, values incorporated in a code description provide rich informative features, potentially leading to an increased predictive performance. When using DescEmb, both the code description $d_{i}$ and the associated numeric values can be embedded with the text encoder $\textit{B}_{\phi}$.

\input{table/table1_AUPRC_eICU}
\input{table/table2_AUPRC_MIMIC}

As shown in Figure \ref{value}, we introduce four different value embedding methods. First, Value Aggregation (VA) stands for aggregating the code description and the numeric values together as text. In this setting, because the BERT tokenizer recognizes each value as a word, it sometimes tokenizes a given value in an unnatural way. For example, a number ‘1351’ can be split into two sub-words ‘13’ and ‘51’, which does not best reflect the underlying meaning of the number. Hence, we additionally propose Digit-Split Value Aggregation (DSVA), where we split all numeric values into each digit first, then aggregate with the code description as text. In this way, a number is always tokenized into single digits, but the model still does not consider the place value. For instance, a number ‘1351’ will be tokenized into ‘1’, ‘3’, ‘5’, ‘1’; the model recognizes the first ‘1’ and the last ‘1’ as the same token even though the first ‘1’ represents the thousandth place and the last ‘1’ represents the first place. To mitigate this misunderstanding, we add learnable Digit Place Embedding (DPE) to every digit token indicating its place value, named Digit-Split Value Aggregation + Digit Place Embedding (DSVA+DPE). No further preprocessing was applied for unit measurements such as percent sign (\%), mg, ml, and so on. This can only be applied to a model exploiting a neural text encoder which can add additional value embeddings for each digit. 
Value Concatenated (VC) embeds description and values separately. Similar to the other embedding methods, code descriptions and units of measurement are embedded through the text encoder, while values are passed through additional Multi-Layer Perceptron (MLP) which yields an embedding vector for the values. These two embeddings are finally concatenated and work as a description representation $\textbf{\textit{z}}_{i}$ for input of the predictive model.
\vspace{-3mm}
\subsection{Model Optimization}
\vspace{-2mm}

Both CodeEmb and DescEmb are used for prediction, therefore we can use any typical prediction loss function $\mathcal{L}$ such as the cross-entropy loss or mean squared error. For DescEmb, training an entire BERT-like text encoder $\textit{B}_{\phi}$ while optimizing predictive model $\textit{P}_{\theta}$ requires a significant amount of time and compute resources, which are often inaccessible by small hospitals. Therefore, we propose the following light-weight DescEmb method, CLS-finetune. The objective functions of each model are shown below.
\vspace{-5mm}

\begin{align}
&\mbox{\textit{Code-based Embedding}}& &\argmin_{\theta, \psi} \mathcal{L}(y, \hat{y}) \label{eq:freeze_opt} \\
&\mbox{\textit{Description-based Embedding}}& &\argmin_{\theta, \phi} \mathcal{L}(y, \hat{y}) \\
&\mbox{\textit{DescEmb CLS-finetune}}& &\argmin_{\theta, \textbf{\textit{z}}_{CLS}} \mathcal{L}(y, \hat{y})
\label{eq:finetune_opt}
\end{align}

CLS-finetune, as written in Eq.~\ref{eq:finetune_opt}, keeps $\phi$ of the text encoder fixed but allows for fine-tuning only the medical event embeddings $\textbf{\textit{z}}_{CLS}$ derived from $\textit{B}$. This can also be seen as initializing the parameters of the embedding layer $\textit{E}_{\psi}$ with the values of $\textbf{\textit{z}}_{CLS}$, instead of initializing with random values. CLS-finetune does not solely rely on $\textit{B}$’s ability to derive medical event embeddings, but allows flexibility for the model to adapt to given prediction task with reasonable computation overhead.

%% file: table/table1_AUPRC_eICU.tex
%

\begin{table*}[ht]
    \caption{\label{eicuAUPRC}\textbf{AUPRC of CodeEmbs and DescEmbs in prediction tasks for eICU}}
    \centering

    \resizebox{0.8\textwidth}{!}{\begin{tabular}{cccccccccc}
    \toprule
    \multirow{2}{*}{}                                    & \multirow{2}{*}{Model}                    & \multicolumn{2}{c}{\multirow{2}{*}{CodeEmb}}        & \multicolumn{6}{c}{DescEmb}                                                                                                                                          \\ \cline{5-10} 
                                                         &                                      & \multicolumn{2}{c}{}                                & \multicolumn{4}{c}{BERT}                                                                            & \multicolumn{2}{c}{RNN}                                        \\ \hline
    \hline
    
    \multicolumn{1}{c|}{Task}                            & \multicolumn{1}{c|}{Value Embedding} & RD                    & \multicolumn{1}{c|}{W2V}    & CLS-FT & FT     & Scr   & \multicolumn{1}{c|}{\begin{tabular}[c]{@{}c@{}}FT \\  + MLM\end{tabular}} & Scr    & \begin{tabular}[c]{@{}c@{}}Scr \\  + MLM\end{tabular} \\ \hline
    \multicolumn{1}{c|}{\multirow{4}{*}{\textbf{Dx}}}    & \multicolumn{1}{c|}{VA}              & 0.447                 & \multicolumn{1}{c|}{0.433}  & 0.501  & 0.574  & 0.547 & \multicolumn{1}{c|}{0.586}                                                & 0.586  & 0.582                                                 \\
    \multicolumn{1}{c|}{}                                & \multicolumn{1}{c|}{DSVA}            & 0.447                 & \multicolumn{1}{c|}{0.433}  & 0.498  & 0.591  & 0.567 & \multicolumn{1}{c|}{0.601}                                                & 0.593  & 0.584                                                 \\
    \multicolumn{1}{c|}{}                                & \multicolumn{1}{c|}{DSVA+DPE}        & \multicolumn{1}{l}{\quad -} & \multicolumn{1}{l|}{\quad -}      & -      & 0.594  & 0.571 & \multicolumn{1}{c|}{0.602}                                                & 0.594  & 0.583                                                 \\
    \multicolumn{1}{c|}{}                                & \multicolumn{1}{c|}{VC}              & 0.562                 & \multicolumn{1}{c|}{0.549}  & 0.557  & 0.562  & 0.546 & \multicolumn{1}{c|}{0.555}                                                & 0.557  & 0.557                                                 \\ \hline
    \multicolumn{1}{c|}{\multirow{4}{*}{\textbf{Mort}}}  & \multicolumn{1}{c|}{VA}              & 0.112                 & \multicolumn{1}{c|}{0.153}  & 0.209  & 0.177  & 0.17  & \multicolumn{1}{c|}{0.216}                                                & 0.237† & 0.271                                                 \\
    \multicolumn{1}{c|}{}                                & \multicolumn{1}{c|}{DSVA}            & 0.112                 & \multicolumn{1}{c|}{0.153}  & 0.209  & 0.223  & 0.215 & \multicolumn{1}{c|}{0.213}                                                & 0.235  & 0.247                                                 \\
    \multicolumn{1}{c|}{}                                & \multicolumn{1}{c|}{DSVA+DPE}        & \multicolumn{1}{l}{\quad -} & \multicolumn{1}{l|}{\quad -}      & -      & 0.224  & 0.213 & \multicolumn{1}{c|}{0.217}                                                & 0.252  & 0.259                                                 \\
    \multicolumn{1}{c|}{}                                & \multicolumn{1}{c|}{VC}              & 0.24†                 & \multicolumn{1}{c|}{0.239†} & 0.238† & 0.23†  & 0.23† & \multicolumn{1}{c|}{0.223}                                                & 0.237† & 0.227†                                                \\ \hline
    \multicolumn{1}{c|}{\multirow{4}{*}{\textbf{LOS$>$3}}} & \multicolumn{1}{c|}{VA}              & 0.47†                 & \multicolumn{1}{c|}{0.439}  & 0.533  & 0.52   & 0.511 & \multicolumn{1}{c|}{0.514}                                                & 0.537  & 0.539                                                 \\
    \multicolumn{1}{c|}{}                                & \multicolumn{1}{c|}{DSVA}            & 0.47†                 & \multicolumn{1}{c|}{0.439}  & 0.529  & 0.53   & 0.538 & \multicolumn{1}{c|}{0.529}                                                & 0.539  & 0.537                                                 \\
    \multicolumn{1}{c|}{}                                & \multicolumn{1}{c|}{DSVA+DPE}        & \multicolumn{1}{l}{\quad-} & \multicolumn{1}{l|}{\quad -}      & -      & 0.536  & 0.537 & \multicolumn{1}{c|}{0.529}                                                & 0.54   & 0.537                                                 \\
    \multicolumn{1}{c|}{}                                & \multicolumn{1}{c|}{VC}              & 0.525                 & \multicolumn{1}{c|}{0.525}  & 0.528  & 0.523  & 0.524 & \multicolumn{1}{c|}{0.523}                                                & 0.526  & 0.53                                                  \\ \hline
    \multicolumn{1}{c|}{\multirow{4}{*}{\textbf{LOS$>$7}}} & \multicolumn{1}{c|}{VA}              & 0.157                 & \multicolumn{1}{c|}{0.184}  & 0.225  & 0.196† & 0.185 & \multicolumn{1}{c|}{0.196}                                                & 0.224  & 0.237                                                 \\
    \multicolumn{1}{c|}{}                                & \multicolumn{1}{c|}{DSVA}            & 0.157                 & \multicolumn{1}{c|}{0.184}  & 0.225  & 0.216  & 0.222 & \multicolumn{1}{c|}{0.221}                                                & 0.227  & 0.233                                                 \\
    \multicolumn{1}{c|}{}                                & \multicolumn{1}{c|}{DSVA+DPE}        & \multicolumn{1}{l}{\quad -} & \multicolumn{1}{l|}{\quad -}      & -      & 0.22   & 0.219 & \multicolumn{1}{c|}{0.221}                                                & 0.231  & 0.234                                                 \\
    \multicolumn{1}{c|}{}                                & \multicolumn{1}{c|}{VC}              & 0.231                 & \multicolumn{1}{c|}{0.228}  & 0.229  & 0.216  & 0.218 & \multicolumn{1}{c|}{0.218}                                                & 0.222  & 0.224                                                 \\ \hline
    \multicolumn{1}{c|}{\multirow{4}{*}{\textbf{ReAdm}}} & \multicolumn{1}{c|}{VA}              & 0.168                 & \multicolumn{1}{c|}{0.15}   & 0.208  & 0.283  & 0.205 & \multicolumn{1}{c|}{0.283}                                                & 0.269  & 0.279                                                 \\
    \multicolumn{1}{c|}{}                                & \multicolumn{1}{c|}{DSVA}            & 0.168                 & \multicolumn{1}{c|}{0.15}   & 0.206  & 0.284  & 0.264 & \multicolumn{1}{c|}{0.29}                                                 & 0.28   & 0.275                                                 \\
    \multicolumn{1}{c|}{}                                & \multicolumn{1}{c|}{DSVA+DPE}        & \multicolumn{1}{l}{\quad -} & \multicolumn{1}{l|}{\quad -}      & -      & 0.289† & 0.263 & \multicolumn{1}{c|}{0.284}                                                & 0.28   & 0.255                                                 \\
    \multicolumn{1}{c|}{}                                & \multicolumn{1}{c|}{VC}              & 0.217†                & \multicolumn{1}{c|}{0.183†} & 0.194  & 0.272  & 0.256 & \multicolumn{1}{c|}{0.267}                                                & 0.277  & 0.276                                                 \\ \hline
    \bottomrule
    \end{tabular}}
        \begin{flushleft}
        {\footnotesize †: standard deviation $>$ 0.02}
        \end{flushleft}

\end{table*}

%% file: table/table2_AUPRC_MIMIC.tex

\begin{table*}[ht]
    \centering
    \caption{\label{mimicauprc} \textbf{AUPRC of CodeEmb and DescEmb in prediction tasks for MIMIC-III.}}
    \resizebox{0.8\textwidth}{!}{\begin{tabular}{cccccccccc}
    \toprule
     \multirow{2}{*}{}                                    & \multirow{2}{*}{}                    & \multicolumn{2}{c}{\multirow{2}{*}{CodeEmb}} & \multicolumn{6}{c}{DescEmb}                                                                                                                      \\ \cline{5-10} 
                                                         &                                      & \multicolumn{2}{c}{}                         & \multicolumn{4}{c}{BERT}                                                        & \multicolumn{2}{c}{RNN}                                        \\ \hline
    \hline
   
    \multicolumn{1}{c|}{Task}                            & \multicolumn{1}{c|}{Value Embedding} & RD          & W2V                            & CLS-FT & FT     & Scr    & \begin{tabular}[c]{@{}c@{}}FT \\  + MLM\end{tabular} & Scr    & \begin{tabular}[c]{@{}c@{}}Scr \\  + MLM\end{tabular} \\ \hline
    \multicolumn{1}{c|}{\multirow{4}{*}{\textbf{Dx}}}    & \multicolumn{1}{c|}{VA}              & 0.726       & \multicolumn{1}{c|}{0.704}     & 0.733  & 0.76   & 0.747  & \multicolumn{1}{c|}{0.767}                           & 0.767  & 0.762                                                 \\
    \multicolumn{1}{c|}{}                                & \multicolumn{1}{c|}{DSVA}            & 0.726       & \multicolumn{1}{c|}{0.704}     & 0.731  & 0.77   & 0.752  & \multicolumn{1}{c|}{0.776}                           & 0.77   & 0.766                                                 \\
    \multicolumn{1}{c|}{}                                & \multicolumn{1}{c|}{DSVA+DPE}        & -           & \multicolumn{1}{c|}{-}         & -      & 0.771  & 0.752  & \multicolumn{1}{c|}{0.764}                           & 0.768  & 0.763                                                 \\
    \multicolumn{1}{c|}{}                                & \multicolumn{1}{c|}{VC}              & 0.757       & \multicolumn{1}{c|}{0.751}     & 0.752  & 0.756  & 0.745  & \multicolumn{1}{c|}{0.75}                            & 0.755  & 0.753                                                 \\ \hline
    \multicolumn{1}{c|}{\multirow{4}{*}{\textbf{Mort}}}  & \multicolumn{1}{c|}{VA}              & 0.228       & \multicolumn{1}{c|}{0.209}     & 0.346  & 0.343† & 0.31   & \multicolumn{1}{c|}{0.38}                            & 0.383  & 0.398                                                 \\
    \multicolumn{1}{c|}{}                                & \multicolumn{1}{c|}{DSVA}            & 0.228       & \multicolumn{1}{c|}{0.209}     & 0.347  & 0.377  & 0.378  & \multicolumn{1}{c|}{0.379}                           & 0.394† & 0.39                                                  \\
    \multicolumn{1}{c|}{}                                & \multicolumn{1}{c|}{DSVA+DPE}        & -           & \multicolumn{1}{c|}{-}         & -      & 0.378  & 0.372  & \multicolumn{1}{c|}{0.383}                           & 0.4    & 0.393                                                 \\
    \multicolumn{1}{c|}{}                                & \multicolumn{1}{c|}{VC}              & 0.313†      & \multicolumn{1}{c|}{0.334}     & 0.339  & 0.336† & 0.335† & \multicolumn{1}{c|}{0.376}                           & 0.344† & 0.338                                                 \\ \hline
    \multicolumn{1}{c|}{\multirow{4}{*}{\textbf{LOS$>$3}}} & \multicolumn{1}{c|}{VA}              & 0.582       & \multicolumn{1}{c|}{0.585}     & 0.608  & 0.616  & 0.601  & \multicolumn{1}{c|}{0.616}                           & 0.624  & 0.63                                                  \\
    \multicolumn{1}{c|}{}                                & \multicolumn{1}{c|}{DSVA}            & 0.582       & \multicolumn{1}{c|}{0.585}     & 0.608  & 0.624  & 0.617  & \multicolumn{1}{c|}{0.619}                           & 0.631  & 0.632                                                 \\
    \multicolumn{1}{c|}{}                                & \multicolumn{1}{c|}{DSVA +DPE}       & -           & \multicolumn{1}{c|}{-}         & -      & 0.624  & 0.616  & \multicolumn{1}{c|}{0.622}                           & 0.634  & 0.628                                                 \\
    \multicolumn{1}{c|}{}                                & \multicolumn{1}{c|}{VC}              & 0.61        & \multicolumn{1}{c|}{0.614}     & 0.616  & 0.61   & 0.614  & \multicolumn{1}{c|}{0.612}                           & 0.622  & 0.622                                                 \\ \hline
    \multicolumn{1}{c|}{\multirow{4}{*}{\textbf{LOS$>$7}}} & \multicolumn{1}{c|}{VA}              & 0.269       & \multicolumn{1}{c|}{0.251}     & 0.346  & 0.338  & 0.325  & \multicolumn{1}{c|}{0.342}                           & 0.349  & 0.349                                                 \\
    \multicolumn{1}{c|}{}                                & \multicolumn{1}{c|}{DSVA}            & 0.269       & \multicolumn{1}{c|}{0.251}     & 0.348  & 0.355  & 0.359  & \multicolumn{1}{c|}{0.356}                           & 0.35   & 0.35                                                  \\
    \multicolumn{1}{c|}{}                                & \multicolumn{1}{c|}{DSVA+DPE}        & -           & \multicolumn{1}{c|}{-}         & -      & 0.36   & 0.359  & \multicolumn{1}{c|}{0.353}                           & 0.352  & 0.353                                                 \\
    \multicolumn{1}{c|}{}                                & \multicolumn{1}{c|}{VC}              & 0.326       & \multicolumn{1}{c|}{0.342}     & 0.346  & 0.341  & 0.339  & \multicolumn{1}{c|}{0.344}                           & 0.347  & 0.352                                                 \\ \hline
    \multicolumn{1}{c|}{\multirow{4}{*}{\textbf{ReAdm}}} & \multicolumn{1}{c|}{VA}              & 0.044       & \multicolumn{1}{c|}{0.043}     & 0.042  & 0.042  & 0.045† & \multicolumn{1}{c|}{0.044}                           & 0.044  & 0.043                                                 \\
    \multicolumn{1}{c|}{}                                & \multicolumn{1}{c|}{DSVA}            & 0.044       & \multicolumn{1}{c|}{0.043}     & 0.041  & 0.043  & 0.046† & \multicolumn{1}{c|}{0.044}                           & 0.045  & 0.044                                                 \\
    \multicolumn{1}{c|}{}                                & \multicolumn{1}{c|}{DSVA+DPE}        & -           & \multicolumn{1}{c|}{-}         & -      & 0.043  & 0.047  & \multicolumn{1}{c|}{0.044}                           & 0.041  & 0.044                                                 \\
    \multicolumn{1}{c|}{}                                & \multicolumn{1}{c|}{VC}              & 0.043       & \multicolumn{1}{c|}{0.043}     & 0.044  & 0.045  & 0.047  & \multicolumn{1}{c|}{0.044}                           & 0.044  & 0.044                                                 \\ \hline
    \bottomrule
    \end{tabular}}
        \begin{flushleft}
        {\footnotesize †: standard deviation $>$ 0.02}
        \end{flushleft}
        
\end{table*}

%% file: tex/4Results.tex
\section{Results}

\subsection{Prediction Performance}

To assess the general efficacy of the DescEmb framework, we evaluate both DescEmbs and CodeEmbs across five medical prediction tasks using two separate datasets. The results are in Table \ref{eicuAUPRC} and Table \ref{mimicauprc}. Value embedding methods are abbreviated as explained in the method section. The results for DSVA + DPE in CodeEmb and CLS-FT are blank since they cannot use Digit Place Embedding. In CodeEmb, ‘RD’ represents a randomly initialized embedding layer while ‘W2V’ represents Word2Vec, a pre-training strategy for CodeEmb embedding layer \citep{mikolov2013efficient}. ‘FT’ stands for fine-tuning where we employ existing pre-trained BERT parameters and fine-tune them for the downstream tasks. ‘Scr’ stands for training from scratch where we do not bring the pre-trained BERT but randomly initialize the model. ‘FT + MLM’ is a model that brings a pre-trained model and conducts Masked Language Modeling (MLM) on our dataset after which it is fine-tuned on downstream tasks. ‘Scr + MLM’ is similar to ‘FT + MLM’ but it does not bring the pre-trained model parameter. We utilize the BERT-Tiny architecture for the BERT-based text encoder because there was no significant performance difference among BERT variants across sizes and pre-training techniques specific to clinical domain corpus; detailed results on this part can be found in Appendix~\ref{supp:domain_res}.

\begin{figure*}[ht]
  \centering
  \includegraphics[width=\linewidth]{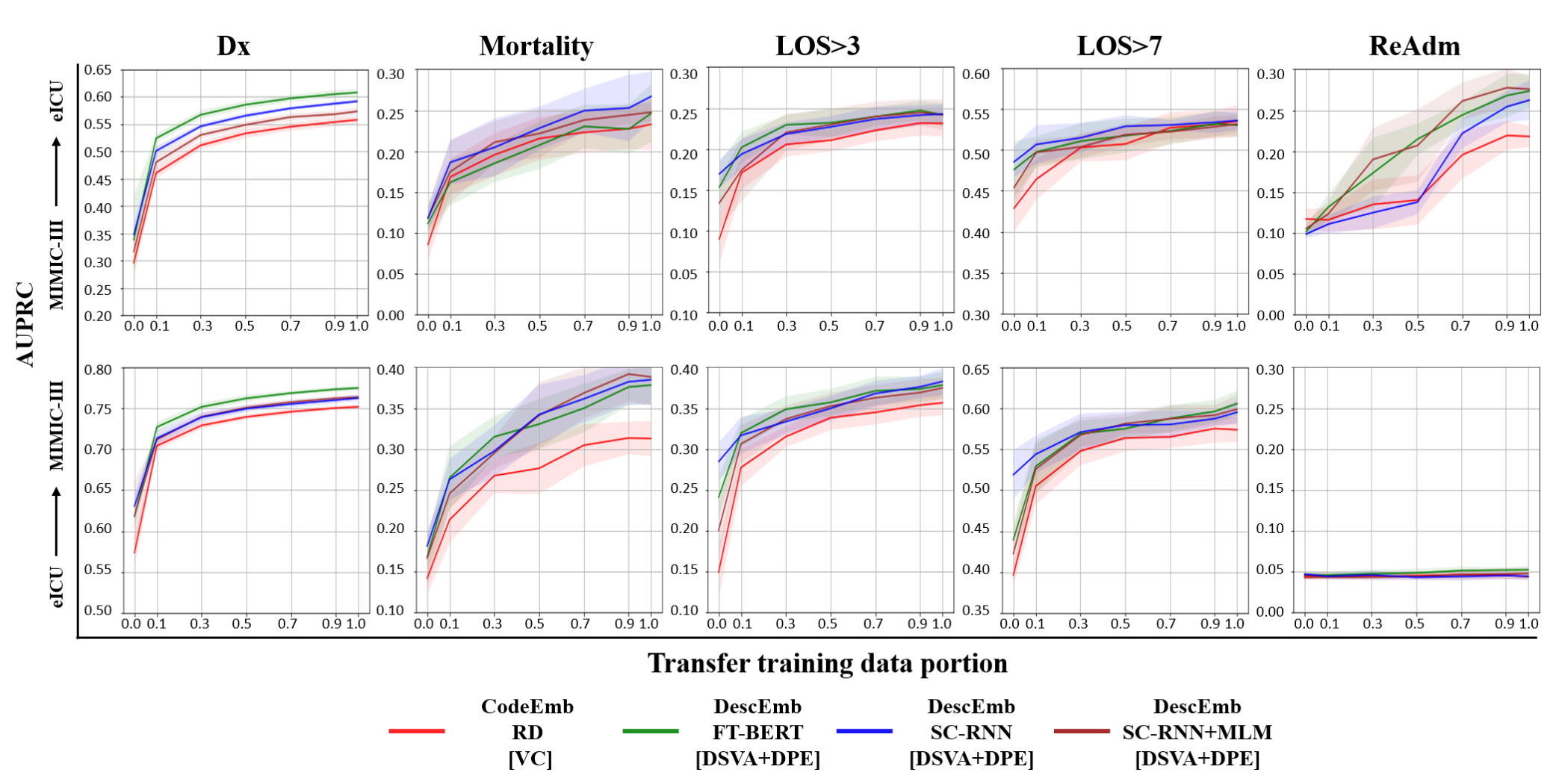}
  \caption{\label{fewshot} \textbf{Transfer learning performance (Top: MIMIC-III to eICU, Bottom: eICU to MIMIC-III).} The X-axis is the portion of the target dataset used for transfer learning, and the Y-axis is the AUPRC at test time on the target dataset. Shading represents the standard deviation from ten seed experiments.}
\end{figure*}

DescEmb models achieve comparable or superior performance to CodeEmb on nearly every task across all value embedding methods at an average of 8$\%$P with 12$\%$P at maximum. Within DescEmb models, BERT-FT generally outperforms BERT-Scratch, verifying the effectiveness of pre-training on massive text corpus. Using the additional Masked Language Modeling (MLM) on our dataset marginally improved performance (+0.3$\%$P AUPRC) for BERT models. We further test the efficacy of MLM in the transfer learning setting below. Of note, a Bi-RNN text encoder generally performs better than BERT-based models. We speculate that, since the maximum lengths of sub-tokens for one code description are 46 and 48 for MIMIC-III and eICU respectively, a simple and light-weighted text encoder model, in this case Bi-RNN, has enough capacity to grasp the features of descriptions. In other words, a large and complex model, in this case BERT, might be an excessively powerful tool to compute refined representations in our setting. We further analyzed factors that influence the most to the prediction, and found that CodeEmb and DescEmb share the same important features under the same task. Detailed results can be found in Appendix~\ref{qualitativeanalysis}.

Note that CLS-finetune in DescEmb, which requires the same amount of compute and time as CodeEmb but initializes the embeddings with the CLS outputs from pre-trained BERT, outperforms CodeEmb in nearly all cases. This demonstrates that there is ground to be gained by adopting description-based embedding compared to the classical code-based embedding. We also pre-train the CodeEmb’s embedding layer in Word2Vec manner to have a fair comparison with the pre-trained DescEmb models. We observe that Word2Vec results are highly unstable, which sometimes underperform 3.4$\%$P at the worst compared to randomly initialized CodeEmb. This result implies that pre-training at code-level is insufficient to fully capture the semantics of each code and sometimes harms the performance. On the other hand, all pre-trained DescEmb models consistently show high performance across all scenarios, verifying the robustness of pre-training at description-level. 

For value embeddings, there is a large discrepancy between CodeEmbs and DescEmbs in Value Aggregation (VA) and Digit-Split Value Aggregation (DSVA) compared to other value embedding methods. We conjecture the underlying reason is that in VA and DSVA, the unique number of codes for CodeEmb explodes since a new code is needed when different values are used.  This raises the curse of dimensionality.  On the contrary, the unique number of sub-tokens used in DescEmb does not change significantly in either setting, resulting in a stable performance. Hence, DescEmb is a suitable model architecture for understanding values because it does not require creating a new code for different values. Value Concatenated (VC) performs the best in CodeEmb. In DescEmb, Digit-Split Value Aggregation with Digit Place Embedding (DSVA+DPE) shows higher performance on the whole than other value embedding methods. It suggests that the model has better numeric understanding since DPE explicitly notifies the model about the place value. For further experiments, we choose CodeEmb RD, FT-BERT, SC-RNN, SC-RNN + MLM, with VC for CodeEmb and DSVA+DPE for the DescEmb models.

\subsection{Zero-Shot Transfer and Few-Shot Transfer}

\input{table/table3_transfer_pooled}

Because DescEmb’s embedding space is determined not by a specific code structure, but rather by the language of the underlying text descriptions, our framework lends itself naturally to transfer learning across all hospitals regardless of their EHR format.  On the other hand, in order to deploy a code-based model on a target dataset with a different code structure than the source dataset, the new code embeddings received by the predictive layer must be randomly initialized, as $\textit{E}_\psi$ is not shared between hospitals. Consequently, CodeEmb’s zero-shot transfer can rely only on the predictive layer parameters whereas DescEmb allows additional flexibility by relying on the $\textit{B}_\phi$ parameters. Here, we transfer one CodeEmb model and three DescEmb models: RD, FT-BERT, SC-RNN, and SC-RNN+MLM trained on the MIMIC-III to eICU dataset and vice versa on zero shot and multiple few shot ratios. For SC-RNN+MLM, we did not conduct additional MLM on the target dataset before the transfer. The results are shown in Figure \ref{fewshot}.

We observe predominantly higher performance of DescEmb over CodeEmb in both the zero-shot and few-shot transfer setting. When this is the case (that is, for all tasks except readmission prediction), DescEmb gains a particular advantage in zero-shot and smaller few-shot ratio transfer learnings, especially for the length-of-stay prediction tasks. This implies that DescEmb can be transferred to different hospitals while retaining its performance even for hospitals with a very small amount of data. We may intuitively understand these results as the uphill battle faced by the CodeEmb to adjust to a completely unfamiliar set of code embeddings—a disadvantage alleviated by a text-based framework and consequently not shared by DescEmb’s predictive layer. Even with a very limited amount of (or no) fine-tuning data from the target dataset, DescEmb can use its knowledge of a prior dataset’s text descriptions to generate effective embeddings at the outset.

\subsection{Pooled Learning with Distinct EHR Formats}

\begin{figure*}[h]
  \centering
  \includegraphics[width=0.8\linewidth]{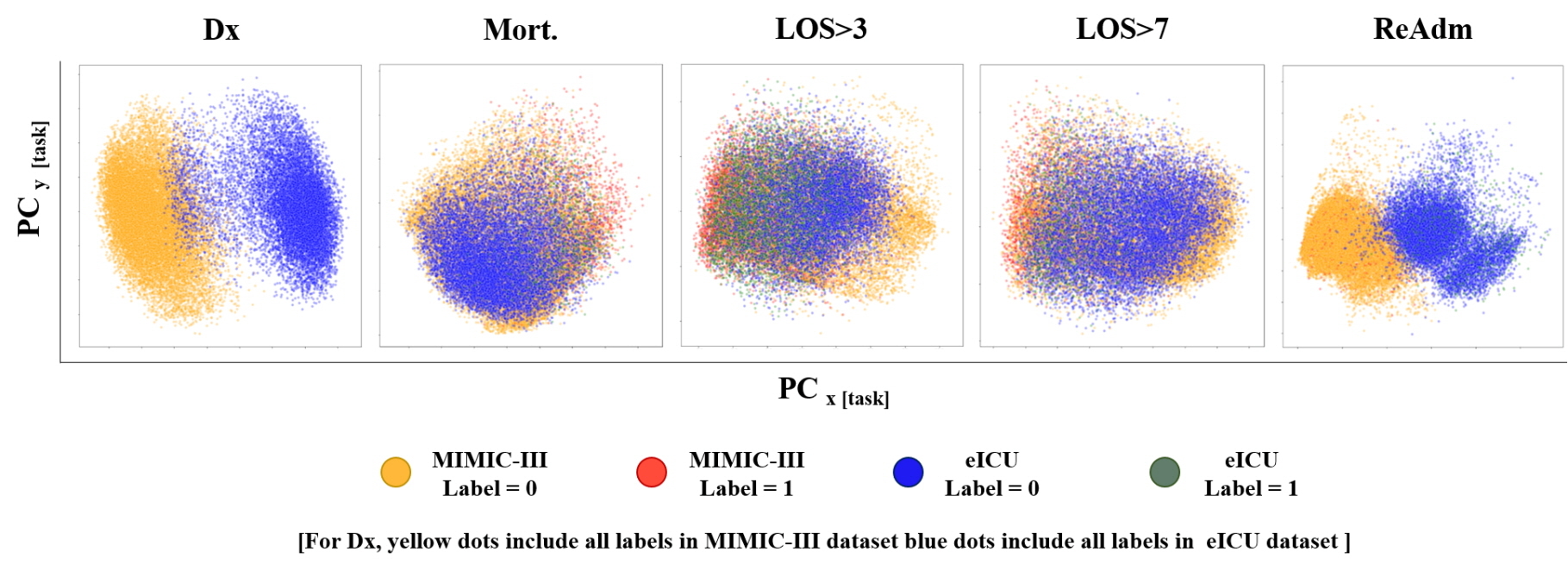}
  \caption{\label{pca} \textbf{PCA visualizations of the ICU representations from the two datasets.} The X-axis and Y-axis correspond to two different principal components. Each dot represents one ICU stay, and the dot color represents the target label for binary classification tasks. For the diagnosis prediction task (Dx), labels could not be succinctly annotated due to its multi-label classification nature. Thus, we distinguish dataset sources by color: yellow being MIMIC-III and blue being eICU.}
\end{figure*}

If we were to deploy a large-scale predictive model in reality, it is more likely that a single central server would pool EHR data from multiple institutions and train a large-scale deep learning model, rather than training many models from individual institutions and performing transfer learning when necessary. Such \textit{pooled learning} presents an opportunity to train a single model to jointly learn information from all institutions. Applying CodeEmb to pooled learning, however, requires either substantial effort to unify code systems (if possible) or relinquishing control over the code vocabulary. Therefore, there is limited benefit of pooled learning for CodeEmb since it consumes a substantial amount of time and labor. Conversely, given that DescEmb is not restricted by specific code structures, pooling datasets does not require any further preprocessing nor extra investment of time and money. 

In order to confirm the efficacy of DescEmb in the pooled learning scenario, we trained both DescEmb and CodeEmb on the pooled training set from both MIMIC-III and eICU, and tested on the individual test set. The results are reported in Table \ref{pooled}, where we compare model performance across various scenarios: train then evaluate on each dataset (“Single MIMIC-III” and “Single eICU”), train on one dataset then fine-tune and evaluate on another dataset (“Transfer eICU→MIMIC-III” and “Transfer MIMIC-III→eICU”), and train on the pooled dataset then evaluate on each dataset (“Pooled MIMIC-III” and “Pooled eICU”). We include both “Single” and “Transfer”, both of which require individual model training on each dataset, to highlight the operational efficiency of pooled learning, which only requires a single model training on the pooled dataset.

Within pooled learning, DescEmb outperformed CodeEmb in all cases (8.9$\%$P at most) except for readmission prediction for MIMIC-III, which indicates that DescEmb is clearly a more suitable framework for pooled learning. Of note, DescEmb’s pooled training showed favorable results compared to the single domain setting as well as transfer learning setting for both MIMIC-III and eICU (more so for eICU which we analyze below). This indicates the efficiency of pooled learning with DescEmb on MIMIC-III and eICU, where only a single model needs to be trained and maintained, instead of training or transferring individual models for each dataset.  Thanks to this efficiency, we believe DescEmb can open new doors for large-scale predictive models in terms of operational cost in finance and time.

\subsection{Representation Distribution and Pooled Learning Advantages}

From Table \ref{pooled}, we can see that eICU generally gained more performance increase than MIMIC-III from both pooled learning and transfer learning. We hypothesize that this comes from the data distribution properties of the two datasets. In order to confirm our hypothesis, we conducted Principal Component Analysis (PCA) on the ICU stay representation vectors obtained from the prediction model (the last hidden layer of the RNN) trained on the pooled dataset. The results in \ref{pca} show that, for some tasks, the eICU representations are distributed inside the MIMIC-III representation distributions, especially in LOS tasks where eICU gained notable performance increase from transfer and pooled learning compared to the single-domain learning.  We deduce that the performance increase comes from learning a more generally distributed dataset, in this case MIMIC-III.

%% file: table/table3_transfer_pooled.tex

\begin{table*}[ht]
    \caption{\label{pooled} \textbf{AUPRC of the models on the five prediction tasks in the three scenarios: single domain learning, transfer learning, pooled learning.}  We compared the AUPRC of code-based embedding model (CodeEmb), pre-trained BERT model (FT-BERT), RNN model (SC-RNN), and RNN model pre-trained on Masked Language Modeling (SC-RNN+MLM).    Based on a t-test, a statistically meaningful increase and decrease against “Single” is marked with boldface and underline, respectively.}
    \centering
    \resizebox{0.8\textwidth}{!}{\begin{tabular}{cc|ccc|ccc}
    \toprule
                                     &                        & \multirow{2}{*}{\begin{tabular}[c]{@{}c@{}}Single\\ MIMIC-III\end{tabular}} & \multirow{2}{*}{\begin{tabular}[c]{@{}c@{}}Transfer\\ eICU \\  → MIMIC-III\end{tabular}} & \multirow{2}{*}{\begin{tabular}[c]{@{}c@{}}Pooled \\  MIMIC-III\end{tabular}} & \multirow{2}{*}{\begin{tabular}[c]{@{}c@{}}Single\\ eICU\end{tabular}} & \multirow{2}{*}{\begin{tabular}[c]{@{}c@{}}Transfer\\ MIMIC-III \\ → eICU\end{tabular}} & \multirow{2}{*}{\begin{tabular}[c]{@{}c@{}}Pooled\\ eICU\end{tabular}} \\
                \multirow{2}{*}{Task}                                & \multirow{2}{*}{Model} &                                                                             &                                                                                          &                                                                               &                                                                        &                                                                                         &                                                                        \\
                                     &                        & \multicolumn{1}{l}{}                                                        & \multicolumn{1}{l}{{\ul }}                                                               & \multicolumn{1}{l|}{}                                                         & \multicolumn{1}{l}{}                                                   & \multicolumn{1}{l}{}                                                                    & \multicolumn{1}{l}{}                                                   \\ \hline
    \hline
    
    \multicolumn{1}{c|}{\multirow{4}{*}{\textbf{Dx}}}    & CodeEmb    & 0.757                                                                       & {\ul{0.752}\text{**}}                                                                   & {0.755}                                                                & 0.562                                                                  & {0.558}                                                                          & {0.563}                                                         \\
    \multicolumn{1}{c|}{}                                & FT-BERT    & 0.771                                                                       & \textbf{0.775*}                                                                          & \textbf{0.777*}                                                               & 0.594                                                                  & \textbf{0.608**}                                                                        & \textbf{0.611*}                                                        \\
    \multicolumn{1}{c|}{}                                & SC-RNN     & 0.768                                                                       & 0.762                                                                                    & \textbf{0.773**}                                                              & 0.594                                                                  & \textbf{0.602**}                                                                        & {0.589}                                                         \\
    \multicolumn{1}{c|}{}                                & SC-RNN+MLM & 0.763                                                                       & 0.76                                                                                     & 0.768                                                                         & 0.583                                                                  & 0.586                                                                                   & \textbf{0.595*}                                                        \\ \hline
    \multicolumn{1}{c|}{\multirow{4}{*}{\textbf{Mort}}}  & CodeEmb    & 0.313                                                                       & 0.313                                                                                    & 0.313                                                                         & 0.24                                                                   & {0.233}                                                                          & {0.247}                                                         \\
    \multicolumn{1}{c|}{}                                & FT-BERT    & 0.378                                                                       & 0.378                                                                                    & 0.376                                                                         & 0.224                                                                  & \textbf{0.246*}                                                                         & \textbf{0.248*}                                                        \\
    \multicolumn{1}{c|}{}                                & SC-RNN     & 0.4                                                                         & 0.385                                                                                    & 0.401                                                                         & 0.252                                                                  & \textbf{0.267*}                                                                         & 0.252                                                                  \\
    \multicolumn{1}{c|}{}                                & SC-RNN+MLM & 0.393                                                                       & 0.383                                                                                    & 0.402                                                                         & 0.259                                                                  & 0.263                                                                                   & {0.253}                                                         \\ \hline
    \multicolumn{1}{c|}{\multirow{4}{*}{\textbf{LOS$>$3}}} & CodeEmb    & 0.61                                                                        & {0.606}                                                                           & 0.611                                                                         & 0.525                                                                  & {0.531}                                                                          & \textbf{0.534*}                                                        \\
    \multicolumn{1}{c|}{}                                & FT-BERT    & 0.624                                                                       & \textbf{0.628*}                                                                          & 0.624                                                                         & 0.536                                                                  & \textbf{0.542*}                                                                         & \textbf{0.549*}                                                        \\
    \multicolumn{1}{c|}{}                                & SC-RNN     & 0.634                                                                       & 0.632                                                                                    & {0.63}                                                                 & 0.54                                                                   & 0.543                                                                                   & \textbf{0.549*}                                                        \\
    \multicolumn{1}{c|}{}                                & SC-RNN+MLM & 0.628                                                                       & 0.627                                                                                    & \textbf{0.638*}                                                               & 0.537                                                                  & 0.541                                                                                   & \textbf{0.548*}                                                        \\ \hline
    \multicolumn{1}{c|}{\multirow{4}{*}{\textbf{LOS$>$7}}} & CodeEmb    & 0.326                                                                       & 0.333                                                                                    & 0.334                                                                         & 0.233                                                                  & {0.235}                                                                          & \textbf{0.239*}                                                        \\
    \multicolumn{1}{c|}{}                                & FT-BERT    & 0.36                                                                        & 0.356                                                                                    & 0.354                                                                         & 0.22                                                                   & \textbf{0.230*}                                                                         & \textbf{0.242**}                                                       \\
    \multicolumn{1}{c|}{}                                & SC-RNN     & 0.352                                                                       & 0.345                                                                                    & 0.35                                                                          & 0.229                                                                  & \textbf{0.236*}                                                                         & \textbf{0.253**}                                                       \\
    \multicolumn{1}{c|}{}                                & SC-RNN+MLM & 0.353                                                                       & 0.342                                                                                    & 0.342                                                                         & 0.234                                                                  & 0.235                                                                                   & \textbf{0.239*}                                                        \\ \hline
    \multicolumn{1}{c|}{\multirow{4}{*}{\textbf{ReAdm}}} & CodeEmb    & 0.043                                                                       & 0.044                                                                                    & 0.049                                                                         & 0.217                                                                  & {0.218}                                                                             & \textbf{0.232*}                                                        \\
    \multicolumn{1}{c|}{}                                & FT-BERT    & 0.043                                                                       & 0.044                                                                                    & 0.051                                                                         & 0.289                                                                  & {\ul{0.274}\text{*}}                                                                            & 0.281                                                                  \\
    \multicolumn{1}{c|}{}                                & SC-RNN     & 0.041                                                                       & 0.045                                                                                    & 0.046                                                                         & 0.28                                                                   & 0.263                                                                                   & {0.279}                                                         \\
    \multicolumn{1}{c|}{}                                & SC-RNN+MLM & 0.044                                                                       & 0.044                                                                                    & 0.044                                                                         & 0.255                                                                  & 0.255                                                                                   & \textbf{0.275*}                                                        \\ \hline
    \bottomrule
    \end{tabular}}
    \begin{flushleft}
    {\footnotesize* : p value $<$ 0.05, \footnotesize** : p value $<$ 0.01}
    \end{flushleft}
\end{table*}

%% file: tex/5Discussion.tex
\vspace{-5mm}
\section{Conclusion}
In this work we introduced a new predictive modeling framework for EHR, namely the description-based embedding (DescEmb), which unifies heterogeneous code systems by deriving the medical code embeddings with a neural text encoder. In a series of experiments with two public EHR datasets and five ICU-based prediction tasks, we demonstrated DescEmb’s outperformance of CodeEmb. We also showed improved zero-shot and few-shot transfer learning performance thanks to the code-agnostic nature of DescEmb. Lastly, we showed that DescEmb provides operational efficiency by enabling us to train a single unified predictive model based on MIMIC-III and eICU, rather than training separate models for each EHR system. We believe this new framework will launch a new discussion around large-scale model training for EHR. Similar to BERT, which has been pre-trained on large text corpus and shown its robustness on text-based tasks, future work includes constructing a large scale EHR pre-trained model through unifying and pooling various hospitals in heterogeneous systems. This pre-trained model can be applied to any time-series EHR dataset without going through laborious pre-processing, which is cost-effective for engineers. Also, incorporating additional modalities such as clinical notes or radiology images can be a key direction for future work.

%% file: tex/Appendix_1.tex
\section{AUPRC Results from Pre-Trained Text Encoders with Different Sizes and Pre-Training Techniques }
\label{supp:domain_res}
\input{table/TableA1-S1.tex}
\input{table/TableA1-S2.tex}

Table \ref{s1} and Table \ref{s2} show AUPRC results differing in the size of pre-trained BERTs (BERT-tiny, BERT-mini, BERT-small, BERT) and in the domain-specific pre-training techniques (Bio-BERT, Bio-Clinical-BERT, Blue-BERT) in eICU and MIMIC-III respectively.  In this experiment, we tested CLS-FT and FT-BERT for verifying the effectiveness of the variants. From the table, there is no consistent performance tendency among different sizes of BERTs and pre-training techniques across tasks and models with very marginal performance differences. Of note, large text encoders generally underperform smaller sizes of BERT. Contrary to our expectation, domain-specialized pre-training techniques rather harm the model performance compared to smaller sizes of BERTs. Overall, the size of the text encoder influences the performance greatly more than how pre-training techniques are modeled. For the experiments in the main paper, we choose BERT-tiny since it generally shows decent performance among other models and it requires less memory and computation time compared to the large models. 

%% file: table/TableA1-S1.tex
\begin{table*}[h]
    \caption{\label{s1}\textbf{Results of BERT variation models on eICU}}
    \centering
    \resizebox{\textwidth}{!}{\begin{tabular}{cc|cccc|ccc}
    \toprule
    Task                                                 & Model   & BERT-tiny & BERT-mini & BERT-small & BERT  & Bio-BERT & Bio-clinical-BERT & Blue-BERT \\ \hline
    \hline
    
    \multicolumn{1}{c|}{\multirow{2}{*}{\textbf{Dx}}}    & CLS-FT  & 0.557     & 0.559     & 0.558      & 0.556 & 0.556    & 0.558             & 0.559     \\
    \multicolumn{1}{c|}{}                                & FT-BERT & 0.594     & 0.595     & 0.595      & 0.591 & 0.59     & 0.593             & 0.591     \\ \hline 
    \multicolumn{1}{c|}{\multirow{2}{*}{\textbf{Mort}}}  & CLS-FT  & 0.238     & 0.242     & 0.233      & 0.228 & 0.231    & 0.228             & 0.228     \\
    \multicolumn{1}{c|}{}                                & FT-BERT & 0.224     & 0.223     & 0.22       & 0.219 & 0.219    & 0.215             & 0.216     \\ \hline 
    \multicolumn{1}{c|}{\multirow{2}{*}{\textbf{LOS>3}}} & CLS-FT  & 0.528     & 0.528     & 0.526      & 0.524 & 0.527    & 0.525             & 0.526     \\
    \multicolumn{1}{c|}{}                                & FT-BERT & 0.536     & 0.527     & 0.523      & 0.523 & 0.522    & 0.528             & 0.526     \\ \hline 
    \multicolumn{1}{c|}{\multirow{2}{*}{\textbf{LOS>7}}} & CLS-FT  & 0.229     & 0.233     & 0.228      & 0.222 & 0.223    & 0.226             & 0.228     \\
    \multicolumn{1}{c|}{}                                & FT-BERT & 0.22      & 0.218     & 0.215      & 0.214 & 0.213    & 0.217             & 0.215     \\ \hline
    \multicolumn{1}{c|}{\multirow{2}{*}{\textbf{ReAdm}}} & CLS-FT  & 0.194     & 0.239     & 0.238      & 0.231 & 0.239    & 0.223             & 0.237     \\
    \multicolumn{1}{c|}{}                                & FT-BERT & 0.289     & 0.283     & 0.278      & 0.276 & 0.277    & 0.281             & 0.275     \\ \hline
    \end{tabular}}
\end{table*}

%% file: table/TableA1-S2.tex
\begin{table*}[h]
    \caption{\label{s2}\textbf{Results of BERT variation models on MIMIC-III}}
    \centering
    \resizebox{\textwidth}{!}{\begin{tabular}{cc|cccc|ccc}
    \toprule
    Task                                                 & Model   & BERT-tiny & BERT-mini & BERT-small & BERT  & Bio-BERT & Bio-clinical-BERT & Blue-BERT \\ \hline
    \hline
    
    \multicolumn{1}{c|}{\multirow{2}{*}{\textbf{Dx}}}    & CLS-FT  & 0.752     & 0.754     & 0.755      & 0.757 & 0.755    & 0.755             & 0.754     \\
    \multicolumn{1}{c|}{}                                & FT-BERT & 0.771     & 0.77      & 0.77       & 0.767 & 0.769    & 0.769             & 0.767     \\ \hline
    \multicolumn{1}{c|}{\multirow{2}{*}{\textbf{Mort}}}  & CLS-FT  & 0.339     & 0.345     & 0.34       & 0.344 & 0.339    & 0.338             & 0.335     \\
    \multicolumn{1}{c|}{}                                & FT-BERT & 0.378     & 0.371     & 0.365      & 0.362 & 0.363    & 0.363             & 0.364     \\ \hline
    \multicolumn{1}{c|}{\multirow{2}{*}{\textbf{LOS>3}}} & CLS-FT  & 0.616     & 0.614     & 0.615      & 0.615 & 0.611    & 0.61              & 0.608     \\
    \multicolumn{1}{c|}{}                                & FT-BERT & 0.624     & 0.623     & 0.623      & 0.621 & 0.626    & 0.622             & 0.62      \\ \hline
    \multicolumn{1}{c|}{\multirow{2}{*}{\textbf{LOS>7}}} & CLS-FT  & 0.346     & 0.344     & 0.341      & 0.343 & 0.344    & 0.344             & 0.338     \\
    \multicolumn{1}{c|}{}                                & FT-BERT & 0.36      & 0.352     & 0.342      & 0.342 & 0.345    & 0.345             & 0.343     \\ \hline
    \multicolumn{1}{c|}{\multirow{2}{*}{\textbf{ReAdm}}} & CLS-FT  & 0.044     & 0.043     & 0.044      & 0.045 & 0.044    & 0.045             & 0.044     \\
    \multicolumn{1}{c|}{}                                & FT-BERT & 0.043     & 0.043     & 0.044      & 0.043 & 0.045    & 0.043             & 0.043     \\ \hline
    \end{tabular}}
\end{table*}

%% file: tex/Appendix_2.tex
\section{PCA Results for Varied Random Seeds}
\begin{figure*}[ht]
  \centering
  \includegraphics[width=0.8\linewidth]{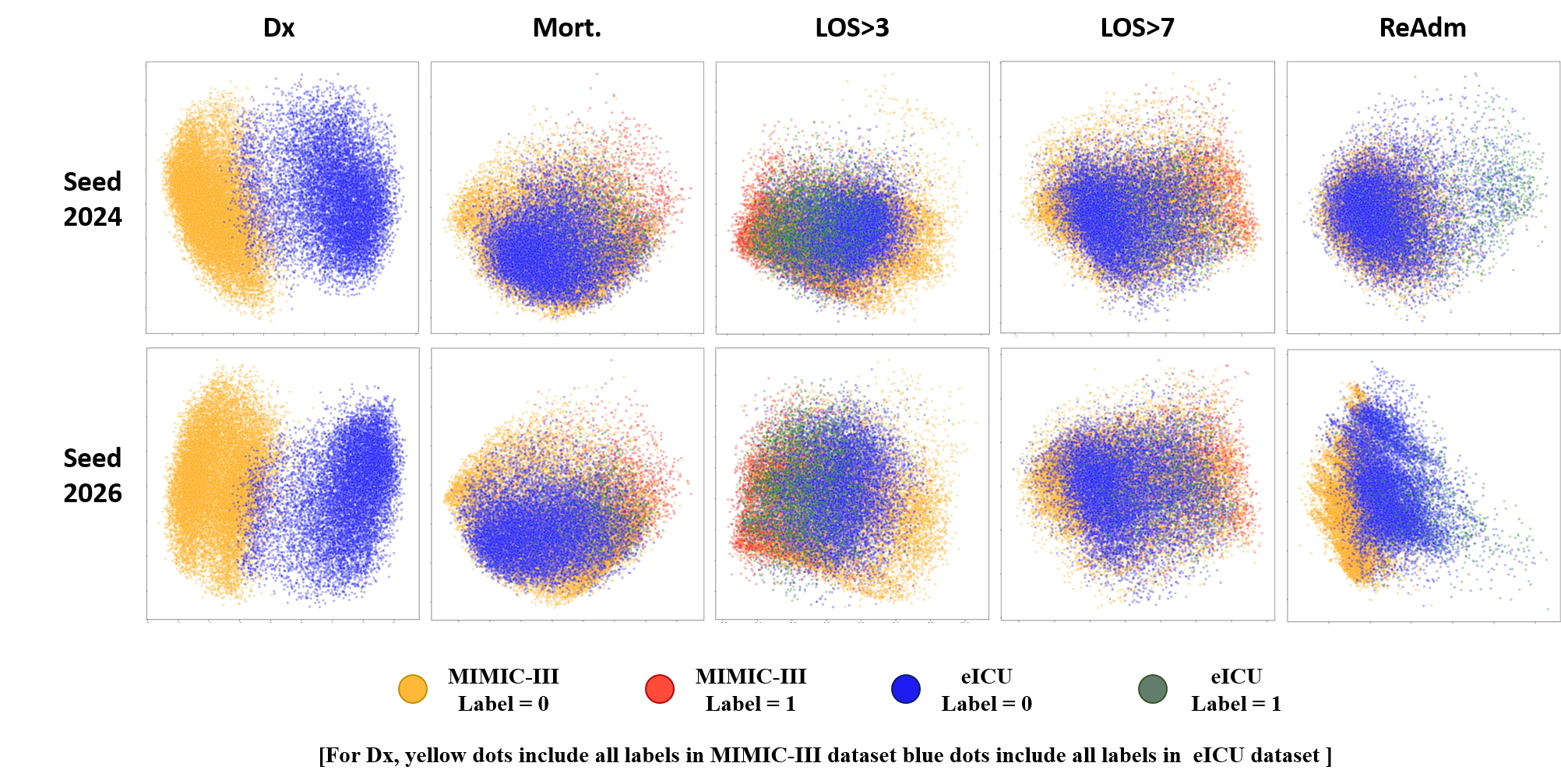}
  \caption{\label{pca_appen} \textbf{PCA visualizations on the ICU Representations for different random seeds.}}
\end{figure*}

We show the PCA results while varying random seeds, which results in a differently split dataset. Figure \ref{pca_appen} shows a similar result to Figure \ref{pca} in the main paper.

%% file: tex/Appendix_3.tex
\section{Detailed preprocessing method and table statistics}
\label{supp:preprocess}
\subsection{Detailed Preprocessing Information}
\label{supp:preprocess_stat}
In the following section we provide further detail about the construction of our datasets. As input for our predictive models, we employ three sources of information (we will further denote source of information as ‘item’ for simplicity)— laboratory, medication, and infusion—simultaneously for each patient. The .csv files corresponding to each item are described in Table \ref{filesource}. Note that when merging MIMIC-III files 'INPUTEVENTS\_MV’ and 'INPUTEVENTS\_CV’, we remove 41 patient histories which straddle the transition between code systems and consequently are included partially in each file.

\input{table/TableA3-S1.tex}

For the sake of comparability, we built patients cohorts from the full MIMIC-III and eICU databases based on the following criteria: (1) Medical ICU (MICU) patients (2) over the age of 18 who (3) remain in the ICU for over 12 hours. We operationalize criterion (1) in MIMIC-III as patients for whom the first care unit is the last care unit and ICU type is MICU (i.e. we exclude patients who have transferred ICUs). For patients with multiple ICU stays, we draw exclusively on the first stay, and we remove any ICU stays with fewer than 5 observed codes. Within each ICU stay, we restrict our sample to the first 150 codes during the first 12 hours of data, and remove codes which occur fewer than 5 times in the entire dataset. Code sequence is determined by the associated time stamp.

\input{table/TableA3-S2.tex}

\subsection{Predictive Task Labels}
\label{supp:preprocess_label}
We predict patient outcomes across five tasks: readmission, mortality, an ICU stay exceeding three or seven days, and diagnosis prediction. The first four are binary classification, the last multi-label. The variable-level criteria to generate these labels is available in Table \ref{labelcriteria}.

In order to generate diagnosis labels for comparison across datasets, we employ the Clinical Classifications Software (CCS) for ICD-9-CM of the Healthcare Cost and Utilization Project \citep{healthcare2016hcup}. We utilize the highest level representation available of ICD9 diagnosis, a common code format across EHR. There are 18 such representations. MIMIC-III and eICU diagnoses represented by ICD9 codes are simply mapped using the CCS classification. eICU ICD10 diagnoses are mapped first to ICD9 codes before to their CCS classification. Finally, for eICU string diagnoses (e.g. Infection ... $|$ ... bacterial ... $|$ ... tuberculosis), we first search the most granular level for a string match with ICD9 before proceeding up the hierarchy for a match.

\input{table/TableA3-S3.tex}

\subsection{Data Statistics}
After preprocessing input data, we found that some patients lack all three items. Consequently, in some cases the item was left out from the patient dataset. For example, some patients have all the items in the code sequence, while others are included without all of them.  In the MIMIC-III and eICU we use, the size of the entire dataset is the same as the union shown in Table \ref{predstats} for each of the source dataset.

\subsection{Hyperparameters}
\label{supp:hyperparameter}
We conducted the hyperparameter searching experiment in CodeEmb and DescEmb on MIMIC-III and eICU. We swept the hyperparameter space within a fixed range, presented below, by grid search.
\begin{itemize}
    \item dropout = [0.1, 0.3, 0.5]
    \item embedding dimension = [128, 256, 512, 768]
    \item hidden dimension = [128, 256, 512]
    \item learning rate = [5e-4, 1e-4, 5e-5, 1e-5]
\end{itemize}
We spent over 72 hours trying to find the best hyperparameter set for each case. We noticed that hyperparameters did not significantly affect the final result. For the experiment’s simplicity, we unified one hyperparameter set for all cases without greatly harming each individual model’s performance. The final set results are dropout of 0.3, embedding dimension and hidden dimension for the predictive model as 128 and 256 respectively, and learning rate of 1e-4.

%% file: table/TableA3-S1.tex
\begin{table}[h]
    \caption{\label{filesource} \textbf{File sources for each dataset}}
    \centering
    \begin{tabular}{l c c }
    \toprule
    Item & Source & Filename \\
    \hline
    \midrule
    Lab &  MIMIC-III  & LABEVENTS.csv  \\
    Lab &  eICU   & lab.csv        \\
    Med &  MIMIC-III  & PRESCRIPTIONS.csv\\
    Med &  eICU   & medication.csv \\
    Inf &  MIMIC-III  & INPUTEVENTS.csv\\
    Inf &  eICU   & infusionDrug.csv \\
    \bottomrule
    \end{tabular}
    
\end{table}

%% file: table/TableA3-S2.tex
\begin{table}[h]
    \caption{\label{predstats} \textbf{Prediction dataset summary statistics}}
    \centering
    \begin{tabular}{l l l }
    \toprule
    Statistic          &  eICU       &  MIMIC-III  \\
    \hline
    \midrule
    $N$ Observations   &  12,818     &    18,536                \\
    $N$ ICU Stays      &  12,818     &    18,536                \\
    $N$ Hospital Adm.  &  12,818     &    18,536                \\
    $N$ Patients       &  12,818     &    18,536                \\
    Mean Seq. Length   &  48.8       &    65.3                  \\
    Median Seq. Length &  43.0       &    57.0                  \\
    $N$ Total Codes    &  625,594    &    1,211,107             \\
    $N$ Unique Codes   &  2,018      &    2,855                 \\
    \bottomrule
    \end{tabular}
\end{table}

%% file: table/TableA3-S3.tex
\begin{table*}[h]
    \caption{\label{labelcriteria} \textbf{Specific label criteria}}
    
    \centering
    \begin{tabular}{l l l}
    \toprule
    Target                  & eICU                                       & MIMIC-III             \\
    \hline
    \midrule
    Readmission             & Count(‘patientUnitStayID') \textgreater 1  & Count(‘ICUSTAY\_ID') \textgreater 1  \\
    Mortality               & ‘unitDischargeStatus'==‘Expired'           & ‘DOD\_HOSP' not null                  \\
    LOS \textgreater 3 Days & ‘unitDischargeOffset' \textgreater 3*24*60 & LOS \textgreater 3                  \\
    LOS \textgreater 7 Days & ‘unitDischargeOffset' \textgreater 7*24*60 & LOS \textgreater 7                  \\
    Diagnosis               & set(‘diagnosisstring’) per 1 ICU           & ICD9\_CODE-LONG\_TITLE     \\
    \bottomrule
    \end{tabular}
\end{table*}

%% file: tex/Appendix_4.tex
\section{Case visualize in description embedding}

 In pooled dataset situation, we conducted PCA on all text descriptions and explored codes that contain `hydrocortisone' with different unit of measurement and dosage, which are colored as red in Figure we attached. The result demonstrates that similar text descriptions with small variants of measurement and dosage are located close to each other. The descriptions for the above points are as follows.
[`hydrocortisone pf iv push', `hydrocortisone na succ. iv', `hydrocortisone po/ng', `hydrocortisone study drug *ind* iv', `hydrocortisone cream 1\% tp', `hydrocortisone na succinate pf iv', `hydrocortisone sod succinate iv', `hydrocortisone po', `hydrocortisone rectal 2.5\% cream pr', `hydrocortisone pf iv'] 
\begin{figure}[h]
  \centering
  \includegraphics[width=0.8\linewidth]{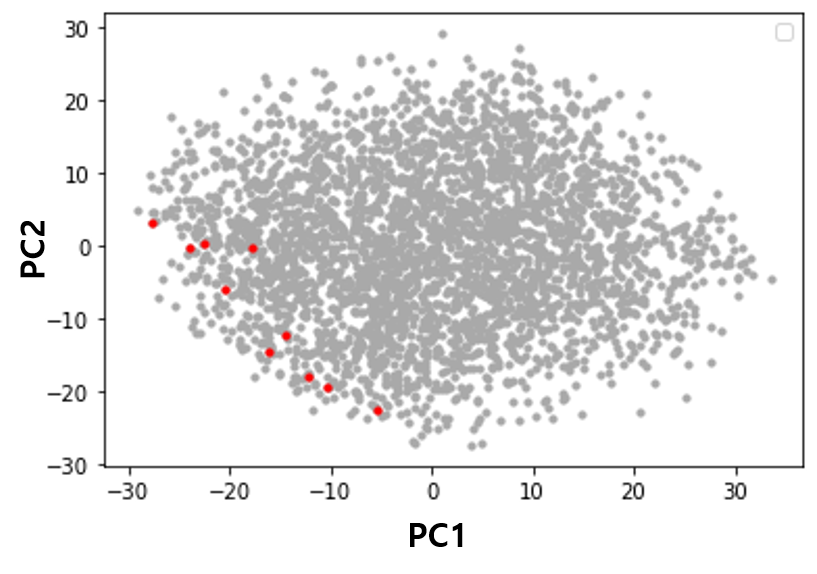}
  \caption{\label{text_pca} \textbf{PCA visualizations on the representation of medication with suffix variation in pooled.}}
\end{figure}

%% file: tex/Appendix_5.tex
\section{Qualitative analysis for features importance in prediction}
\label{qualitativeanalysis}
\begin{figure*}[ht]
  \centering
  \includegraphics[width=1.0\linewidth]{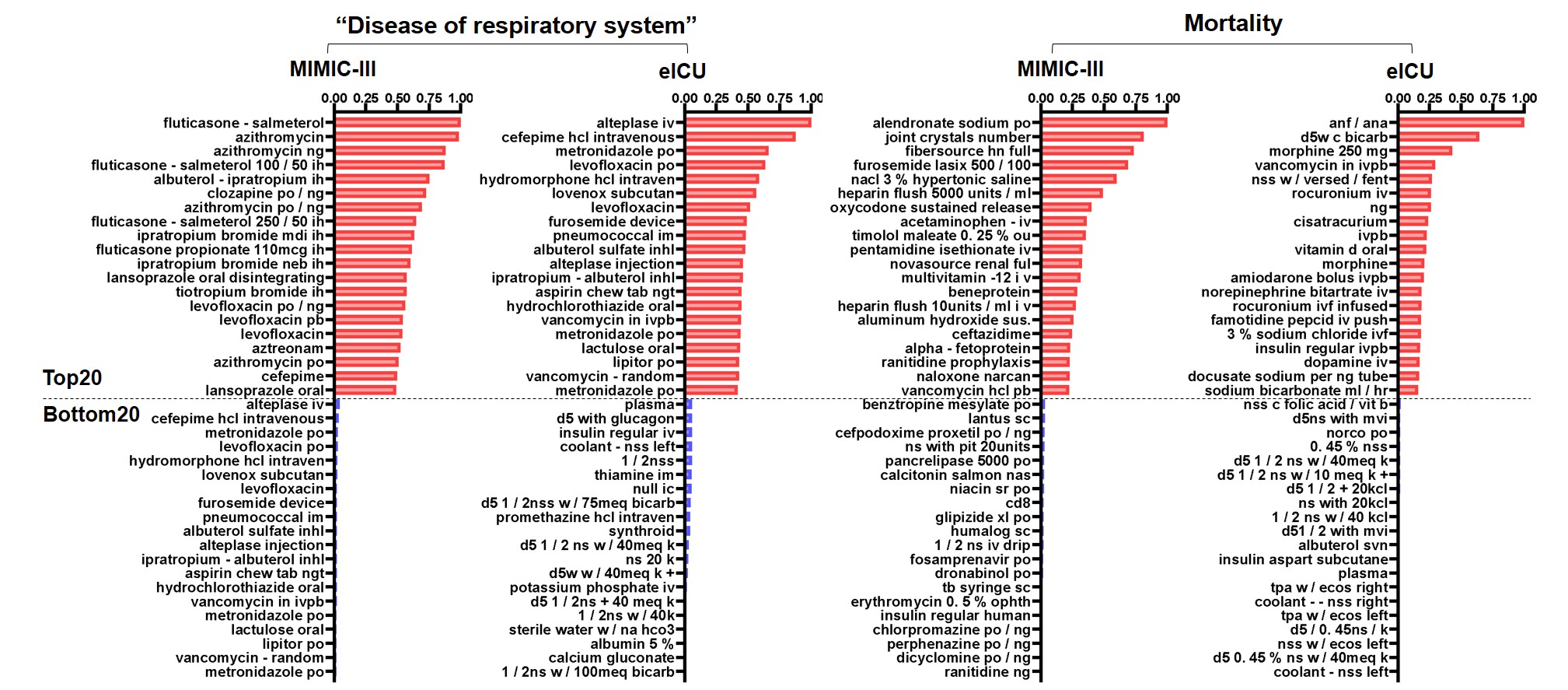}
  \caption{\label{features} \textbf{Top and bottom 20 features for “Disease of respiratory system” and “mortality” prediction model with DescEmb Scr-RNN.}}
\end{figure*}
\subsection{Top and bottom features for diagnosis and mortality prediction in DescEmb Scr-RNN model}

 We used the gradient of backpropagation of each feature; the larger the gradient, the more impactful the feature. Figure \ref{features} shows the 20 features most and least contributing to the prediction for ‘Disease for respiratory system’ in diagnosis prediction (left two) and ‘mortality’ (right two) on MIMIC-III and eICU using DescEmb RNN-Scratch (further named as RNN-Scr). For ‘Disease for respiratory system’, we trained a new model to conduct the binary classification (respiratory system vs not respiratory system) instead of multi-label classification which was presented in the paper to better analyze the contributing features. In ‘Disease for respiratory system’, most influential factors were related to medication whereas in ‘Mortality’, most factors were a combination of medication and lab values. In ‘Mortality’, we noticed that medicines often prescribed to those close to death exist in the top 20.

\subsection{Comparison of top features between different models in diagnosis prediction tasks}
Tables \ref{features_nervous}, \ref{features_circulatory}, \ref{features_respiratory}, \ref{features_digestive} show the most influential factors for five tasks (Disease of the nervous system and sense organs, Disease of circulatory system, Disease of digestive system, Disease of respiratory system) for DescEmb RNN-Scr, Bert-FT, and CodeEmb on MIMIC-III and eICU. In ‘Disease of nervous system’, it mainly contains medications that are consumed in the department of neurology. In ‘Disease of Circulatory’, there are diuretics (‘furosemide’) and machines that measure blood pressure. There also are blood transfusions, antithrombotic, antibiotic, and other heart medicines for those who had operations. For ‘Disease of Respiratory’, most of the factors are antibiotics utilized on pneumonia and lung injury. In ‘Disease of Digestive’, most factors are medicines prescribed for digestive and gastrointestinal inflammation. 
In conclusion, the top most influential features under each task are reasonable and are highly related to the given task. Also, most features are shared across models (DescEmb RNN-Scr, BERT-FT, and CodeEmb) and datasets (MIMIC-III and eICU). 

\input{table/TableA5-S1.tex}
\input{table/TableA5-S2.tex}
\input{table/TableA5-S3.tex}
\input{table/TableA5-S4.tex}

%% file: table/TableA5-S1.tex
\begin{table*}[h]
    \caption{\label{features_nervous} \textbf{Top features for “Disease of the nervous system and sense organs” prediction models with DescEmb and CodeEmb}}
    \centering
     \resizebox{1.0\textwidth}{!}{\begin{tabular}{lccc}
    \toprule
    \multirow{2}{*}{MIMIC-III} & \multicolumn{2}{c}{DescEmb} & CodeEmb
    \\
    & RNN-Scr & Bert-FT & RD
    \\\hline
    \midrule
    \\
&	 latanoprost 0. 005 \% ophth. soln. ou     	&	 artificial tear ointment ou                 	&	 timolol maleate 0. 5 \% both eyes           	\\
&	 latanoprost 0. 005 \% ophth. soln. left eye    	&	 brimonidine tartrate 0. 15 \% ophth. os      	&	 carbamazepine                  	\\
&	 pyridostigmine bromide po            	&	 latanoprost 0. 005 \% ophth. soln. ou     	&	 carbidopa - levodopa cr 50 - 200 po       	\\
&	 erythromycin 0. 5 \% ophth oint od       	&	 topiramate topamax po / ng  	&	 ropinirole	\\
&	 latanoprost 0. 005 \% ophth. soln. right eye    	&	 carbamazepine      	&	 topiramate topamax po               	\\
&	 carbidopa - levodopa cr 50 - 200 po       	&	 sumatriptan succinate sc              	&	 carbamazepine po / ng              	\\
&	 brimonidine tartrate 0. 15 \% ophth. ou       	&	 erythromycin 0. 5 \% ophth oint ou        	&	 ciprofloxacin 0. 3 \% ophth soln both eyes     	\\
&	 gabapentin po                 	&	 ciprofloxacin 0. 3 \% ophth soln both eyes     	&	 sumatriptan succinate sc              	\\
&	 phenobarbital                 	&	 timolol maleate 0. 25 \% both eyes           	&	 oxcarbazepine po               	\\
&	 prazosin po                 	&	 gabapentin po                 	&	 erythromycin 0. 5 \% ophth oint ou        	\\
&	 oxcarbazepine po               	&	 prazosin po                 	&	 losartan potassium po / ng               	\\
&	 latanoprost 0. 005 \% ophth. soln. both eyes    	&	 oxcarbazepine po               	&	 phenobarbital                 	\\
&	 brimonidine tartrate 0. 15 \% ophth. both eyes      	&	 ropinirole	&	 lidocaine 1 \% id                 	\\
&	 carbamide peroxide 6. 5 \% ad            	&	 latanoprost 0. 005 \% ophth. soln. right eye    	&	 valproic acid                 	\\
&	 artificial tears preserv. free both eyes              	&	 latanoprost 0. 005 \% ophth. soln. both eyes    	&	 timolol maleate 0. 25 \% both eyes           	\\
&	 topiramate po         	&	 brimonidine tartrate 0. 15 \% ophth. ou       	&	 phenobarbital po               	\\
&	 hydromorphone - hp ivpca             	&	 pyridostigmine bromide po            	&	 ciprofloxacin 0. 3 \% ophth soln both eyes     	\\
&	 carbamazepine                  	&	 carbamide peroxide 6. 5 \% ad            	&	 artificial tear ointment ou                 	\\
&	 topiramate topamax po / ng  	&	 carbidopa - levodopa cr 50 - 200 po       	&	 topiramate topamax po / ng  	\\
&	 ropinirole	&	 clozapine po                 	&	 fluvoxamine maleate po	\\
    \\\hline
    \multirow{2}{*}{eICU} & & &
    \\
    & & &
    \\\hline
    \bottomrule
    \\
    &	 levetiracetam po                      	&	 levetiracetam po                      	&	 phenytoin                        	\\
&	 phenytoin                        	&	 phenytoin                        	&	 lactulose oral                        	\\
&	 lactulose oral                        	&	 levetiracetam ivpb                    	&	 levetiracetam po                      	\\
&	 naloxone hcl intraven                   	&	 levetiracetam po                      	&	 phenytoin                        	\\
&	 xanax po                        	&	 phenytoin                        	&	 zolpidem tartrate oral                     	\\
&	 phenytoin                        	&	 lactulose oral                        	&	 chlorhexidine gluconate mouth rinse                	\\
&	 lorazepam                         	&	 xanax po                        	&	 levetiracetam po                      	\\
&	 lactulose oral                        	&	 naloxone hcl intraven                   	&	 levetiracetam ivpb                    	\\
&	 zolpidem tartrate oral                     	&	 colace po                         	&	 hydromorphone hcl intraven                   	\\
&	 seroquel oral                         	&	 levetiracetam po                      	&	 phenylephrine mcg / kg / min                 	\\
&	 naloxone hcl intraven                   	&	 seroquel oral                         	&	 naloxone hcl intraven                   	\\
&	 levetiracetam po                      	&	 restoril oral                         	&	 seroquel oral                         	\\
&	 levetiracetam ivpb                    	&	 hydromorphone hcl intraven                   	&	restoril oral                         	\\
&	 phenylephrine mcg / kg / min                 	&	 hydrochlorothiazide oral                    	&	thiamine po	\\
&	 restoril oral                         	&	 naloxone hcl intraven                   	&	 hydrochlorothiazide oral                    	\\
&	 hydromorphone hcl intraven                   	&	 phenylephrine mcg / kg / min                 	&	 lorazepam                         	\\
&	 thiamine po                        	&	 lorazepam                         	&	midazolam versed iv	\\
&	 anf / ana                        	&	 chlorhexidine gluconate mt                  	&	mupirocin top	\\
&	 metoprolol tartrate per g tube                   	&	 amlodipine	&	 anf / ana                        	\\
&	 phenylephrine 	&	 metoclopramide	&	 amlodipine	\\
    \\\hline
    \end{tabular}}
    
\end{table*}

%% file: table/TableA5-S2.tex
\begin{table*}[h]
    \caption{\label{features_circulatory} \textbf{Top features for "Disease of the circulatory system" prediction models with DescEmb and CodeEmb}}
    \centering
     \resizebox{1.0\textwidth}{!}{\begin{tabular}{lccc}
    \toprule
    \multirow{2}{*}{MIMIC-III} & \multicolumn{2}{c}{DescEmb} & CodeEmb
    \\
    & RNN-Scr & Bert-FT & RD
    \\\hline
    \midrule
    \\
    &	 prazosin po                 	&	 prazosin po                 	&	 prazosin po                 	\\
&	 500n / s 40meq k +              	&	 buprenorphine - naloxone 8mg - 2mg sl    	&	 enoxaparin lovenox                	\\
&	 xigris                    	&	 enoxaparin lovenox                	&	 enema                     	\\
&	 enema                     	&	 howell - jolly bodies                 	&	 reticulocyte count automated                	\\
&	 enoxaparin lovenox                	&	 500n / s 40meq k +              	&	 ethacrynate sodium iv  	\\
&	 wright giemsa                 	&	 wright giemsa                 	&	 fibersource hn full                 	\\
&	 leucovorin calcium iv                	&	 reticulocyte count automated                	&	 enoxaparin lovenox                	\\
&	 wright giemsa                 	&	 enema                     	&	 levothyroxine sodium ng	\\
&	 howell - jolly bodies                 	&	 thrombosis	&	 leucovorin calcium iv                	\\
&	 ethacrynate sodium iv  	&	 insulin - humalog     	&	 gray top hold plasma                   	\\
&	 mexiletine po	&	 cd23                    	&	 carafate sucralfate               	\\
&	 buprenorphine - naloxone 8mg - 2mg sl   	&	 mexiletine po	&	 xigris                     	\\
&	 reticulocyte count automated                	&	 ethacrynate sodium iv  	&	 wright giemsa	\\
&	 enema 	&	 fibersource hn full                 	&	 thrombosis	\\
&	 golytely                   	&	 enoxaparin lovenox                	&	 carafate sucralfate               	\\
&	 thrombosis	&	 thrombosis	&	 nutren pulmonary	\\
&	 iv piggyback                  	&	 gray top hold plasma                   	&	 amikacin	\\
&	 gray top hold plasma                   	&	 buprenorphine - naloxone 8mg - 2mg sl   	&	 macrophage                    	\\
&	 carafate sucralfate               	&	 leucovorin calcium iv                	&	 mexiletine po	\\
&	 insulin - humalog     	&	 macrophage                    	&	 buprenorphine - naloxone 8mg - 2mg sl   	\\

    \\\hline
    \multirow{2}{*}{eICU} & & &
    \\
    & & &
    \\\hline
    \bottomrule
    \\
    &	 hydrochlorothiazide po                   	&	 hydrochlorothiazide po                   	&	 lipitor po                         	\\
&	 furosemide lasix intravenous     	&	 hydrochlorothiazide oral                    	&	 hydrochlorothiazide po                   	\\
&	 hydrochlorothiazide oral                    	&	 lipitor po                         	&	 coreg 	\\
&	 furosemide device                        	&	 furosemide lasix intravenous     	&	 amiodarone bolus ivpb                  	\\
&	 lipitor po                         	&	 furosemide device                        	&	 digoxin oral                         	\\
&	 lovenox subcutan                       	&	 alteplase iv                        	&	 hydrochlorothiazide oral                    	\\
&	 alteplase iv                        	&	 bumetanide 	&	 furosemide lasix intravenous     	\\
&	 coreg 	&	 coreg 	&	 furosemide device                        	\\
&	 bumetanide 	&	 lovenox subcutan                       	&	 alteplase iv                        	\\
&	 catheter                         	&	 digoxin oral                         	&	 bumetanide 	\\
&	 digoxin oral                         	&	 amiodarone bolus ivpb                  	&	 lovenox subcutan                       	\\
&	 potassium phosphate dibasic iv                      	&	 phenylephrine mcg / kg / min                 	&	 amiodarone bolus ivpb                  	\\
&	 metoprolol tartrate per g tube                   	&	 wbcs in body fluid                       	&	 wbcs in body fluid                       	\\
&	 amiodarone bolus ivpb                  	&	 potassium phosphate dibasic iv                      	&	 metronidazole po	\\
&	 cefepime hcl intravenous                  	&	 digoxin oral                         	&	 digoxin oral                         	\\
&	 lanoxin oral	&	 furosemide device                        	&	 lanoxin oral	\\
&	 atorvastatin calcium per ng tube   	&	 lanoxin oral	&	 tylenol ng tube                       	\\
&	 tylenol ng tube                       	&	 tylenol ng tube                       	&	 potassium phosphate dibasic iv	\\
&	 cetirizine oral                        	&	 catheter                         	&	 atorvastatin calcium per ng tube   	\\
&	 phenylephrine mcg / kg / min                 	&	 atorvastatin calcium per ng tube   	&	 cefepime hcl intravenous	\\

    \\\hline
    \end{tabular}}
    
\end{table*}

%% file: table/TableA5-S3.tex
\begin{table*}[h]
    \caption{\label{features_respiratory} \textbf{Top features for “Disease of the respiratory system” prediction models with DescEmb and CodeEmb}}
    \centering
     \resizebox{1.0\textwidth}{!}{\begin{tabular}{lccc}
    \toprule
    \multirow{2}{*}{MIMIC-III} & \multicolumn{2}{c}{DescEmb} & CodeEmb
    \\
    & RNN-Scr & Bert-FT & RD
    \\\hline
    \midrule
    \\
&	 fluticasone – salmeterol	&	 fluticasone - salmeterol diskus 100 / 50 ih	&	 albuterol - ipratropium ih	\\
&	 azithromycin	&	 albuterol - ipratropium ih	&	 fluticasone - salmeterol 100 / 50 ih	\\
&	 azithromycin ng	&	 azithromycin	&	 aztreonam	\\
&	 fluticasone - salmeterol 100 / 50 ih	&	 azithromycin ng	&	 azithromycin po	\\
&	 albuterol - ipratropium ih	&	 levofloxacin po / ng	&	 levofloxacin pb	\\
&	 clozapine po / ng	&	 tiotropium bromide ih	&	 cefepime	\\
&	 azithromycin po / ng	&	 azithromycin po / ng	&	 tiotropium bromide ih	\\
&	 fluticasone - salmeterol 250 / 50 ih	&	 azithromycin po	&	 fluticasone - salmeterol 100 / 50 ih	\\
&	 ipratropium bromide mdi ih	&	 fluticasone - salmeterol 100 / 50 ih	&	 lansoprazole oral disintegrating tab	\\
&	 fluticasone propionate 110mcg ih	&	 aztreonam	&	 levofloxacin po / ng 	\\
&	 ipratropium bromide neb ih	&	 clozapine po / ng	&	 metronidazole po	\\
&	 lansoprazole oral disintegrating	&	 azithromycin po	&	 levofloxacin	\\
&	 tiotropium bromide ih	&	 ipratropium bromide neb ih	&	 azithromycin	\\
&	 levofloxacin po / ng	&	 cefepime	&	 azithromycin ng	\\
&	 levofloxacin pb	&	 fluticasone - salmeterol 250 / 50 ih	&	 nutren pulmonary full                   	\\
&	 levofloxacin	&	 levofloxacin pb	&	 ipratropium bromide mdi ih	\\
&	 aztreonam	&	 lansoprazole oral disintegrating tab 	&	 fluticasone propionate 110mcg ih	\\
&	 azithromycin po	&	 levofloxacin	&	 lansoprazole oral suspension ng	\\
&	 cefepime	&	 carafate sucralfate               	&	 tiotropium bromide ih	\\
&	 lansoprazole oral	&	 nutren pulmonary full                   	&	 ipratropium bromide neb ih	\\

    \\\hline
    \multirow{2}{*}{eICU} & & &
    \\
    & & &
    \\\hline
    \bottomrule
    \\
    &	 alteplase iv	&	 cefepime hcl intravenous  	&	 alteplase iv                        	\\
&	 cefepime hcl intravenous 	&	 furosemide device                        	&	 cefepime hcl intravenous  	\\
&	 metronidazole po	&	 levofloxacin po                     	&	 pneumococcal im                      	\\
&	 levofloxacin po	&	 alteplase iv                        	&	 ipratropium - albuterol inhl                 	\\
&	 hydromorphone hcl intraven	&	 lovenox subcutan                       	&	 lovenox subcutan                       	\\
&	 lovenox subcutan	&	 hydrochlorothiazide oral                    	&	 levofloxacin	\\
&	 levofloxacin	&	 metronidazole po	&	 levofloxacin po                     	\\
&	 furosemide device	&	 aspirin chew tab ngt                    	&	 hydromorphone hcl intraven                   	\\
&	 pneumococcal im	&	 hydromorphone hcl intraven                   	&	 vancomycin in ivpb	\\
&	 albuterol sulfate inhl	&	 pneumococcal im                      	&	 zolpidem tartrate oral	\\
&	 alteplase injection	&	 vancomycin in ivpb	&	 hydrochlorothiazide oral                    	\\
&	 ipratropium - albuterol inhl	&	 albuterol sulfate inhl                    	&	 hydromorphone hcl intraven                   	\\
&	 aspirin chew tab ngt	&	 ipratropium - albuterol inhl                 	&	 aspirin chew tab ngt                    	\\
&	 hydrochlorothiazide oral	&	 hydrochlorothiazide oral                    	&	 metronidazole po                      	\\
&	 vancomycin in ivpb 	&	 restoril oral                         	&	 cefepime hcl intravenous	\\
&	 metronidazole po	&	 lipitor po                         	&	 alteplase injection	\\
&	 lactulose oral	&	 metronidazole po                      	&	 bumetanide 	\\
&	 lipitor po	&	 zolpidem tartrate oral                     	&	 pneumococcal im                      	\\
&	 vancomycin – random	&	 alteplase injection	&	 potassium chloride device	\\
&	 metronidazole po	&	 vancomycin - random	&	 chlorhexidine periogard swish / spit                	\\
    
    \\\hline
    \end{tabular}}
    
\end{table*}

%% file: table/TableA5-S4.tex
\begin{table*}[h]
    \caption{\label{features_digestive} \textbf{Top features for “Disease of the digestive system” prediction models with DescEmb and CodeEmb}}
    \centering
     \resizebox{1.0\textwidth}{!}{\begin{tabular}{lccc}
    \toprule
    \multirow{2}{*}{MIMIC-III} & \multicolumn{2}{c}{DescEmb} & CodeEmb
    \\
    & RNN-Scr & Bert-FT & RD
    \\\hline
    \midrule
    \\
    &	 omeprazole prilosec              	&	 omeprazole prilosec              	&	 promethazine hcl po              	\\
&	 ranitidine prophylaxis                	&	 ranitidine prophylaxis                	&	 misoprostol po               	\\
&	 triamcinolone acetonide 0. 1 \% cream tp       	&	 carafate sucralfate               	&	 omeprazole prilosec              	\\
&	 carafate sucralfate               	&	 protonix mg / hr                	&	 carafate sucralfate	\\
&	 promethazine hcl po              	&	 promethazine hcl po              	&	 lansoprazole prevacid	\\
&	 dialysate in                   	&	 criticare hn                   	&	 protonix mg / hr                	\\
&	 bismuth subsalicylate po             	&	 bismuth subsalicylate po             	&	 ranitidine prophylaxis                	\\
&	 isosource 1. 5 full               	&	 deliver 2. 0                   	&	 pd fluid in                   	\\
&	 beneprotein                   	&	 probalance full                   	&	 prazosin po                 	\\
&	 nutren pulmonary full                   	&	 nutren pulmonary                    	&	 isosource 1. 5 full               	\\
&	 criticare hn                   	&	 beneprotein                   	&	 nutren 2. 0 full                 	\\
&	 deliver 2. 0                   	&	 misoprostol po               	&	 gray top hold plasma	\\
&	 lupus anticoagulant 	&	 prazosin po                 	&	 probalance full	\\
&	 lansoprazole prevacid               	&	 ranitidine prophylaxis                	&	 fibersource hn full	\\
&	 probalance full                   	&	 pd fluid in                   	&	 nutren pulmonary	\\
&	 peptamen 1. 5 full               	&	 nutren 2. 0 full                 	&	 peptamen 1. 5 full               	\\
&	 misoprostol po               	&	 isosource 1. 5 full               	&	 wright giemsa	\\
&	 protonix mg / hr                	&	 dialysate in                   	&	 pyrimethamine desensitization po            	\\
&	 phenytoin free    	&	 triamcinolone acetonide 0. 1 \% cream tp       	&	 criticare hn                   	\\
&	 wright giemsa 	&	 peptamen 1. 5 full               	&	 lidocaine 5\% ointment tp	\\

    \\\hline
    \multirow{2}{*}{eICU} & & &
    \\
    & & &
    \\\hline
    \bottomrule
    \\
    &	 famotidine per g tube                      	&	 famotidine per g tube                      	&	 colace po                         	\\
&	 docusate sodium per ng tube                      	&	 famotidine iv                       	&	 levofloxacin po                     	\\
&	 colace po                         	&	 docusate sodium per ng tube                      	&	 famotidine per g tube                      	\\
&	 docusate sodium colace capsule po                    	&	 colace po                         	&	 famotidine iv                       	\\
&	 famotidine iv                       	&	 docusate sodium colace capsule po                    	&	 docusate sodium colace capsule po                    	\\
&	 levofloxacin po                     	&	 prolactin                         	&	 docusate sodium per ng tube                      	\\
&	 hydromorphone hcl intraven                   	&	 levofloxacin po                     	&	 chlorhexidine gluconate mouth rinse                	\\
&	 thiamine po                        	&	 lactulose oral                        	&	 metronidazole ivpb                    	\\
&	 prolactin                         	&	 hydromorphone hcl intraven                   	&	 prolactin                         	\\
&	 chlorhexidine gluconate mouth rinse                	&	 chlorhexidine gluconate mouth rinse                	&	 albumin human iv                        	\\
&	 lactulose oral                        	&	 albumin human iv                        	&	 hydromorphone hcl intraven                   	\\
&	 cryoprecipitate     	&	 thiamine po                        	&	 d5  0.45\% ns 	\\
&	 albumin human iv                        	&	 metronidazole ivpb                    	&	 thiamine po                        	\\
&	 metronidazole ivpb                    	&	 lactulose oral                        	&	 vancomycin hcl in dextrose iv        	\\
&	 vancomycin hcl in dextrose iv        	&	 d5. 45ns 40kcl                    	&	 lactulose oral                        	\\
&	 potassium chloride device                          	&	 pantoprazole iv                       	&	 octreotide	\\
&	 sterile water w / 3 amps bicarb                 	&	 flagyl iv                         	&	 bumetanide 	\\
&	 potassium phosphate dibasic iv                      	&	 lanoxin oral                         	&	 cryoprecipitate     	\\
&	 zolpidem tartrate oral                     	&	 cryoprecipitate     	&	 potassium chloride device	\\
&	 tirofiban     	&	 sterile water w / 3 amps bicarb                 	&	 potassium phosphate dibasic iv	\\

    \\\hline
    \end{tabular}}
    
\end{table*}